\documentclass[runningheads]{llncs}

\usepackage{eccv}

\usepackage{eccvabbrv}

\usepackage{graphicx}
\usepackage{booktabs}

\usepackage{multirow}
\usepackage{rotating}   
\usepackage{xcolor}
\usepackage{adjustbox}
\usepackage{subcaption}

\newcommand{\best}[1]{\textcolor{red}{#1}}
\newcommand{\second}[1]{\textcolor{blue}{#1}}

\newcommand{\zdeg}{z_{\text{deg}}}

\newcommand{\zclean}{z_{\text{clean}}}

\usepackage[accsupp]{axessibility}  

\usepackage[pagebackref,breaklinks,colorlinks,citecolor=eccvblue]{hyperref}
\usepackage{hyperref}

\usepackage{orcidlink}

\begin{document}

\title{Your Pre-trained Diffusion Model Secretly Knows Restoration} 


\author{Sudarshan Rajagopalan \and
Vishal M. Patel}


\authorrunning{S.~Rajagopalan, V.M.~Patel}

\institute{Johns Hopkins University, Baltimore MD 21218, USA \\
\email{\{sambasa2,vpatel36\}@jhu.edu}}

\maketitle

\begin{abstract}

Pre-trained diffusion models have enabled significant advancements in All-in-One Restoration (AiOR), offering improved perceptual quality and generalization. However, diffusion-based restoration methods primarily rely on fine-tuning or Control-Net style modules to leverage the pre-trained diffusion model's priors for AiOR. In this work, we show that these pre-trained diffusion models inherently possess restoration behavior, which can be unlocked by directly learning prompt embeddings at the output of the text encoder. Interestingly, this behavior is largely inaccessible through text prompts and text-token embedding optimization. Furthermore, we observe that naive prompt learning is unstable because the forward noising process using degraded images is misaligned with  the reverse sampling trajectory. To resolve this, we train prompts within a diffusion bridge formulation that aligns training and inference dynamics, enforcing a coherent denoising path from noisy degraded states to clean images. Building on these insights, we introduce our lightweight learned prompts on the pre-trained WAN video model and FLUX image models, converting them into high-performing restoration models. Extensive experiments demonstrate that our approach achieves competitive performance and generalization across diverse degradations, while avoiding fine-tuning and restoration-specific control modules. Project page: \url{https://sudraj2002.github.io/yptpage/}.

\end{abstract}

\section{Introduction}
\label{sec: intro}

All-in-One Restoration (AiOR) aims to restore images or videos corrupted by various degradations such as haze, rain, snow, motion blur, and low light, using a single unified model. Recent AiOR methods increasingly leverage large pre-trained text-to-image or video (T2IV) latent diffusion models (LDMs) such as Stable Diffusion~\cite{stablediff,sdxl}, Lumina~\cite{lumina,gao2024lumin-t2x}, and FLUX~\cite{flux}. Trained on massive data, these models possess strong priors over the natural image distribution and can generate high-quality content, making them attractive backbones for robust and perceptually pleasing restoration~\cite{autodir,restorevar,pixwizard,diffplugin}. However, most existing approaches harness these priors primarily by adapting the diffusion backbone via fine-tuning~\cite{autodir,unildiff,pixwizard} or by introducing ControlNet-style conditioning modules~\cite{unicorn,diffrestorer}. While effective, these approaches typically require substantial training and can be overly sensitive to the training distribution which can weaken the strong pre-trained priors, limiting robustness under real-world or complex degradations. Alternatively, plug-and-play methods~\cite{pnpflow,pnp2} avoid training but often depend on explicit degradation operators, which are difficult to specify for complex scenarios.

\begin{figure*}[t]
  \centering

  \begin{minipage}[t]{0.24\textwidth}\centering\small Input \end{minipage}\hfill
  \begin{minipage}[t]{0.24\textwidth}\centering\small P2P+NTI\end{minipage}\hfill
  \begin{minipage}[t]{0.24\textwidth}\centering\small SDEdit\end{minipage}\hfill
  \begin{minipage}[t]{0.24\textwidth}\centering\small Ours \end{minipage}

  \begin{subfigure}[t]{0.24\textwidth}
    \centering
    \includegraphics[height=0.8\linewidth, width=\linewidth]{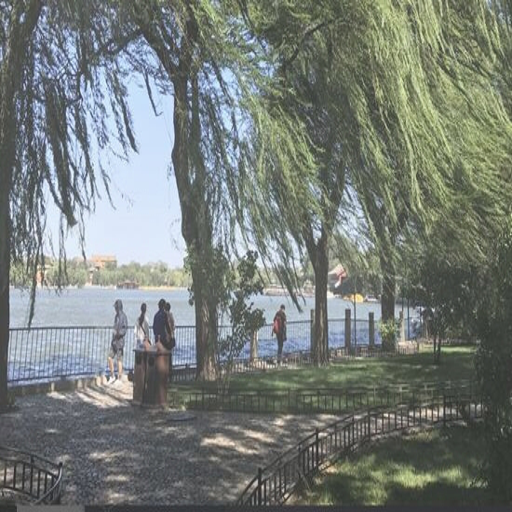}
  \end{subfigure}\hfill
  \begin{subfigure}[t]{0.24\textwidth}
    \centering
    \includegraphics[height=0.8\linewidth, width=\linewidth]{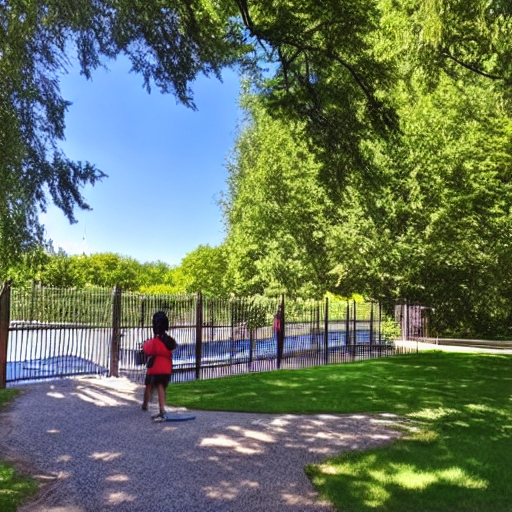}
  \end{subfigure}\hfill
  \begin{subfigure}[t]{0.24\textwidth}
    \centering
    \includegraphics[height=0.8\linewidth, width=\linewidth]{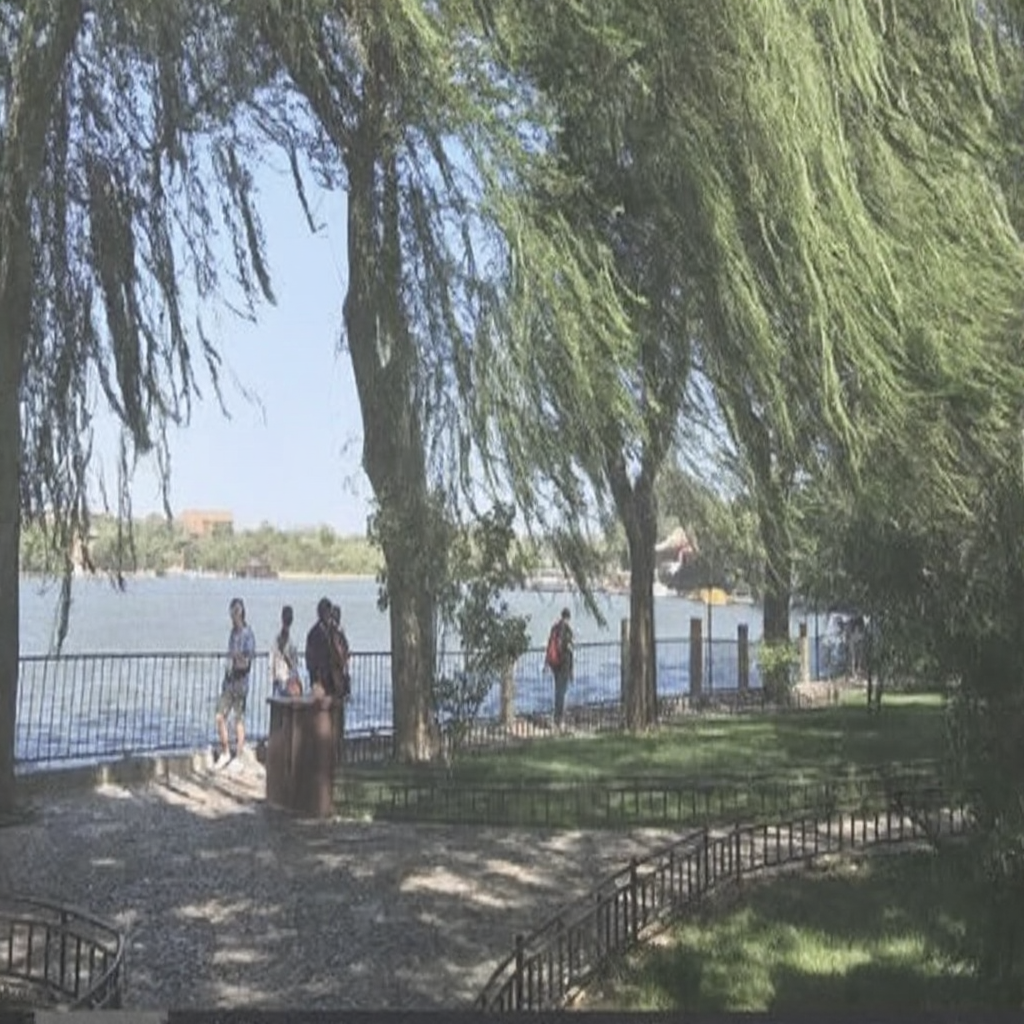}
  \end{subfigure}\hfill
  \begin{subfigure}[t]{0.24\textwidth}
    \centering
    \includegraphics[height=0.8\linewidth, width=\linewidth]{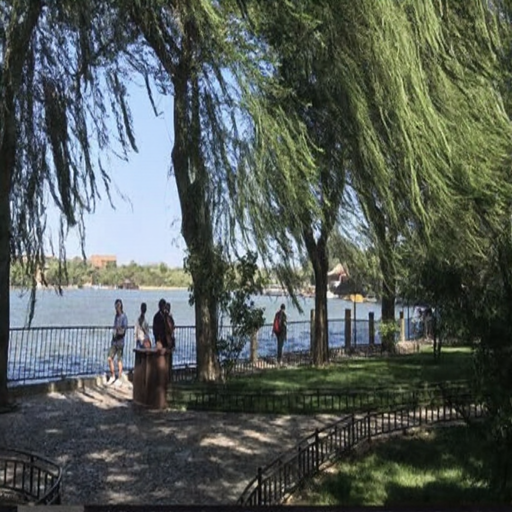}
  \end{subfigure}

  \caption{Popular editing based approaches such as SDEdit~\cite{sdedit}, and Prompt-to-prompt~\cite{p2p} with Null-text inversion~\cite{nti} (P2P + NTI) work well for high-level editing but perform poorly for restoration tasks, in this case dehazing.}
  \label{fig: edit_motiv}
\end{figure*}

In parallel, pre-trained T2IV models have enabled remarkable progress in image editing, where the sampling trajectory is \emph{steered} through language while keeping the diffusion backbone frozen. This paradigm is appealing and versatile because it preserves the model's original priors and achieves strong high-level semantic edits. However, directly transferring such editing strategies to low-level restoration tasks is largely ineffective. Inversion-based editing methods~\cite{p2p,nti} alter image details, while partial-noising approaches such as SDEdit~\cite{sdedit} preserve the input structure but, the natural-language guidance (e.g., ``remove haze''), often fails to remove degradations from intermediate noisy degraded states and converges back to the degraded input with noise suppressed (see Fig.~\ref{fig: edit_motiv}). This raises a key question: \emph{Do large diffusion model priors contain restoration-relevant knowledge, and if so, can we access them through conditioning mechanisms?}

\begin{figure*}[t]
\centering

\begin{minipage}[t]{0.5\textwidth}
  \vspace{0pt} 
  \centering
  \includegraphics[height=4.5cm,width=\linewidth,keepaspectratio]{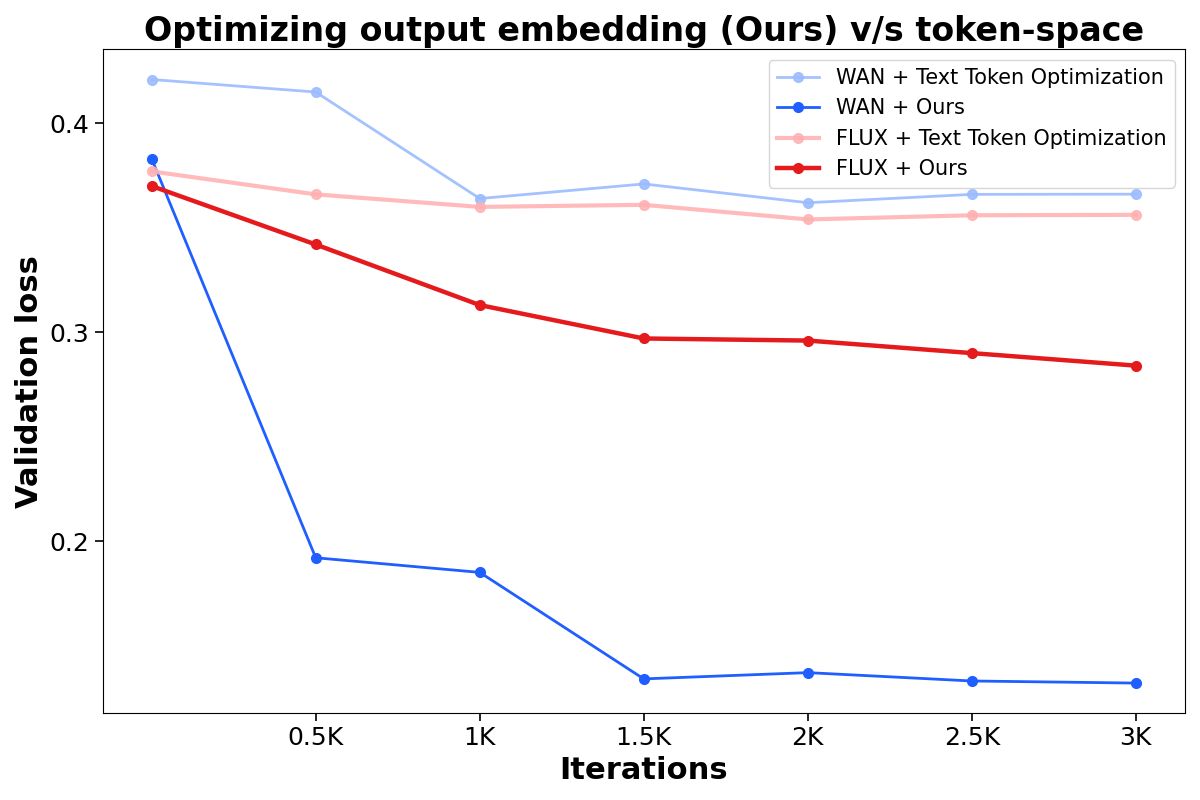}
\end{minipage}\hfill
\begin{minipage}[t]{0.46\textwidth}
  \vspace{0pt} 
  \centering
  \setlength{\tabcolsep}{1.5pt}
  \renewcommand{\arraystretch}{0.9} 
  \begin{tabular}{cc}
    \includegraphics[height=2.6cm,width=0.49\linewidth,keepaspectratio]{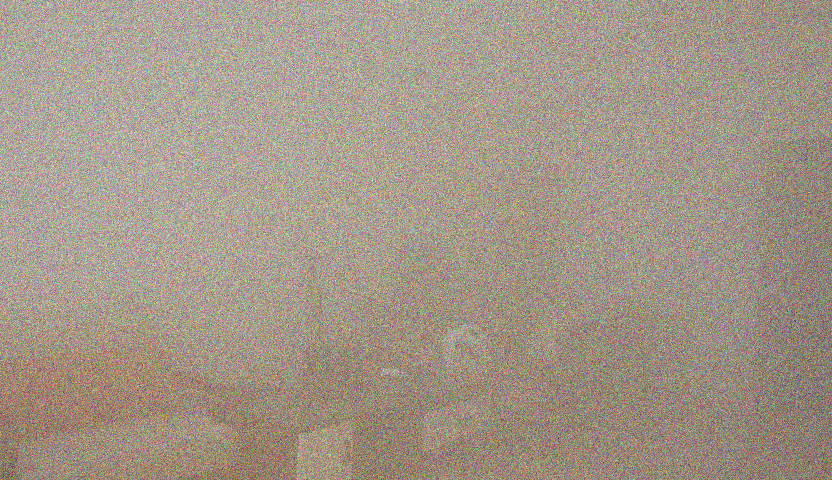} &
    \includegraphics[height=2.6cm,width=0.49\linewidth,keepaspectratio]{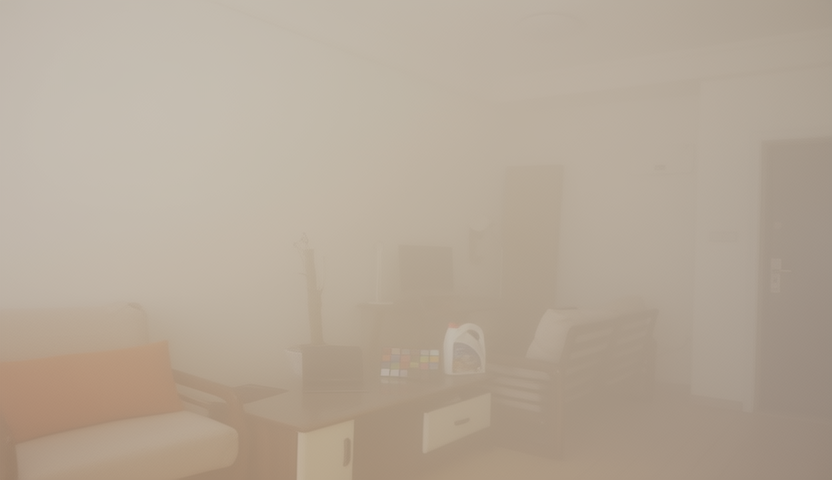} \\
    {\scriptsize Noisy degraded input} & {\scriptsize Text prompt} \\
    \includegraphics[height=2.6cm,width=0.49\linewidth,keepaspectratio]{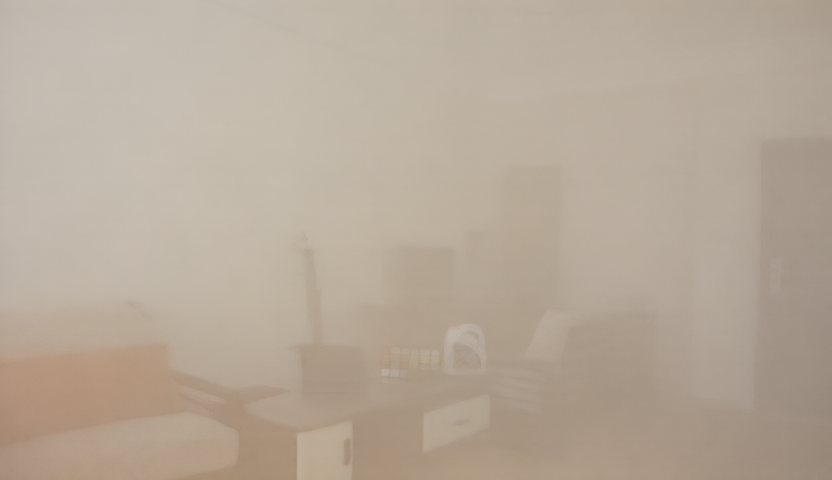} &
    \includegraphics[height=2.6cm,width=0.49\linewidth,keepaspectratio]{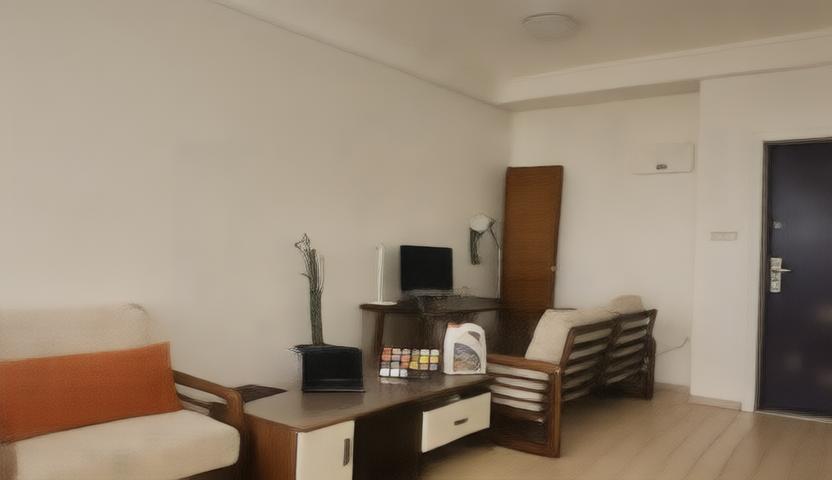} \\
    {\scriptsize Token optimization} & {\scriptsize Ours (embedding)} \\
  \end{tabular}
\end{minipage}

\caption{Text token-space prompting is ineffective for restoration: 
Even with optimized token prompts (textual inversion/prompt tuning), the model tends to denoise without removing degradations, whereas embedding-space optimization enables restoration from the same noisy degraded input.}
\label{fig: text_motiv}
\end{figure*}

In this paper, we investigate this question and reveal that the bottleneck is not the frozen diffusion backbone but the \emph{conditioning representation}. Standard token-space prompts, whether obtained through natural language or optimized as learned tokens (e.g., textual inversion/prompt tuning), fail to elicit restoration behavior (see Fig.~\ref{fig: text_motiv}). In contrast, directly optimizing the conditioning \emph{in embedding space} (i.e., at the output of the text encoder) unlocks strong restoration performance, i.e. the frozen diffusion model consistently maps noisy degraded inputs toward clean targets, suggesting that restoration knowledge exists in the pre-trained diffusion prior but is not accessible through the text tokens. A second challenge is \emph{trajectory mismatch}. Naively training prompts using a forward process of the form $\mathbf{x}_t = \alpha(t)\mathbf{x}_{\text{deg}} + \sigma(t)\boldsymbol{\epsilon}$ leads to a difference in the distributions of $x_t$ during training and inference, which affects prompt performance. To address this issue, we propose to train prompts within a \emph{diffusion-bridge} formulation that aligns trajectories during training and inference-time sampling. We introduce our framework on the WAN 1.3B~\cite{wan} video diffusion model and the FLUX 12B~\cite{flux} image diffusion model, achieving competitive performance against state-of-the-art restoration methods across diverse degradations with significantly fewer trainable parameters. 

To summarize our contributions are as follows:
\begin{itemize}
    \item We show that natural-language prompts and token-space embedding optimization are ineffective for eliciting restoration behavior from pre-trained text-to-image/video diffusion backbones. We then reveal that directly learning the conditioning signal in the text-encoder \emph{output} space unlocks strong restoration while keeping the diffusion backbone frozen.

    \item We identify a train-test trajectory mismatch that makes naive prompt learning unstable, and propose training prompts on bridge-defined intermediate states (with a bridge-compatible reparameterization for WAN/FLUX) to align optimization and sampling dynamics.

    \item We instantiate our framework on the WAN and FLUX models, and show that they achieve competitive restoration performance to the state-of-the-art and strong generalization with only lightweight learned prompts.

\end{itemize}

\section{Related Works}
\label{sec: related}
\subsection{All-in-One Restoration}
\label{subsec: related_aior}

AiOR aims to restore images or videos affected by diverse degradations using a single unified model. Early works such as All-in-One~\cite{nas} and TransWeather~\cite{transw} explored unified architectures and training strategies. Subsequent methods improved performance by introducing learnable prompts such as in PromptIR~\cite{promptir} and visual in-context cues~\cite{awracle}. More recent instruction-guided approaches incorporate language to select restoration behavior, e.g., InstructIR~\cite{instructir} and DFPIR~\cite{dfpir}. Other approaches such as GenDeg~\cite{gendeg} and FoundIR~\cite{foundir} investigate data scaling effects for AiOR. ViWS-Net~\cite{viwsnet} proposed a transformer-based approach for video all-weather removal while AverNet~\cite{avernet} proposed a diffusion-based method for video AiOR. With the rise of large pretrained diffusion models, many AiOR approaches exploit their generative priors to improve perceptual quality and robustness~\cite{autodir,diffplugin,restorevar}. Existing diffusion-based AiOR methods typically leverage these priors through two main paradigms: (i) Backbone adaptation: methods such as AutoDIR~\cite{autodir}, UniLDiff~\cite{unildiff}, and PixWizard~\cite{pixwizard} adapt pretrained diffusion backbones (e.g., Stable Diffusion~\cite{stablediff}) using full or parameter-efficient fine-tuning to follow restoration-based language instructions. While effective, this requires substantial training and can bias the model toward the training data distribution, potentially reducing robustness under complex real-world corruptions. (ii) Learned conditioning modules: another line of work keeps the diffusion backbone frozen but trains ControlNet-style branches or conditioning networks to inject task- or degradation-dependent signals, as in UNICORN~\cite{unicorn} and Diff-Restorer~\cite{diffrestorer}. These modules directly modify intermediate features of the frozen model, which can weaken the model's pretrained priors if the conditioning module is not robust. Plug-and-play methods~\cite{pnpflow,pnp2} perform inference-time guidance using pretrained diffusion priors, but typically depend on explicit degradation operators or optimization procedures, which are difficult to specify for complex or unknown degradations. Unlike the above approaches, we learn compact prompts in the native conditioning pathway of the frozen diffusion model which unlocks their potential for restoration tasks.

\subsection{Pre-trained Diffusion Models for Editing}
\label{subsec: related_dm}

Pre-trained diffusion models have been widely explored for editing tasks, thanks to their strong generative prior. A common paradigm includes inversion-based methods that aim to accurately recover a noise representation that reconstructs the input under the diffusion model~\cite{bracklimitless,cao2023masactrl,nti}. Subsequently, structural or language-driven controls are applied during denoising to realize the desired edits~\cite{p2p,pnpinv,tumanyan2023plug}. While effective for semantic manipulation, these mechanisms often fail to preserve fine-grained fidelity (e.g., background texture, identity, or small structures) due to imperfect inversion and attention interventions which could be particularly problematic for restoration. SDEdit~\cite{sdedit} avoids exact inversion by partially noising the input and denoising from that intermediate state, offering a tunable trade-off between fidelity and the model’s generative prior via the chosen noise level. However, we observe that simply replacing the editing instruction with a restoration instruction (e.g., “remove haze”, “deblur”) performs poorly for restoration as the model tends to suppress noise without removing degradations. Motivated by this gap, we investigate how to enable an SDEdit-like restoration procedure by learning compact prompts that steer denoising toward clean targets while keeping the diffusion backbone frozen.

\section{Proposed Method}
\label{sec: proposed}

In this section, we explain the core innovations of our method along with important design choices.

\begin{figure}[t]
    \centering
    \includegraphics[width=0.9\linewidth]{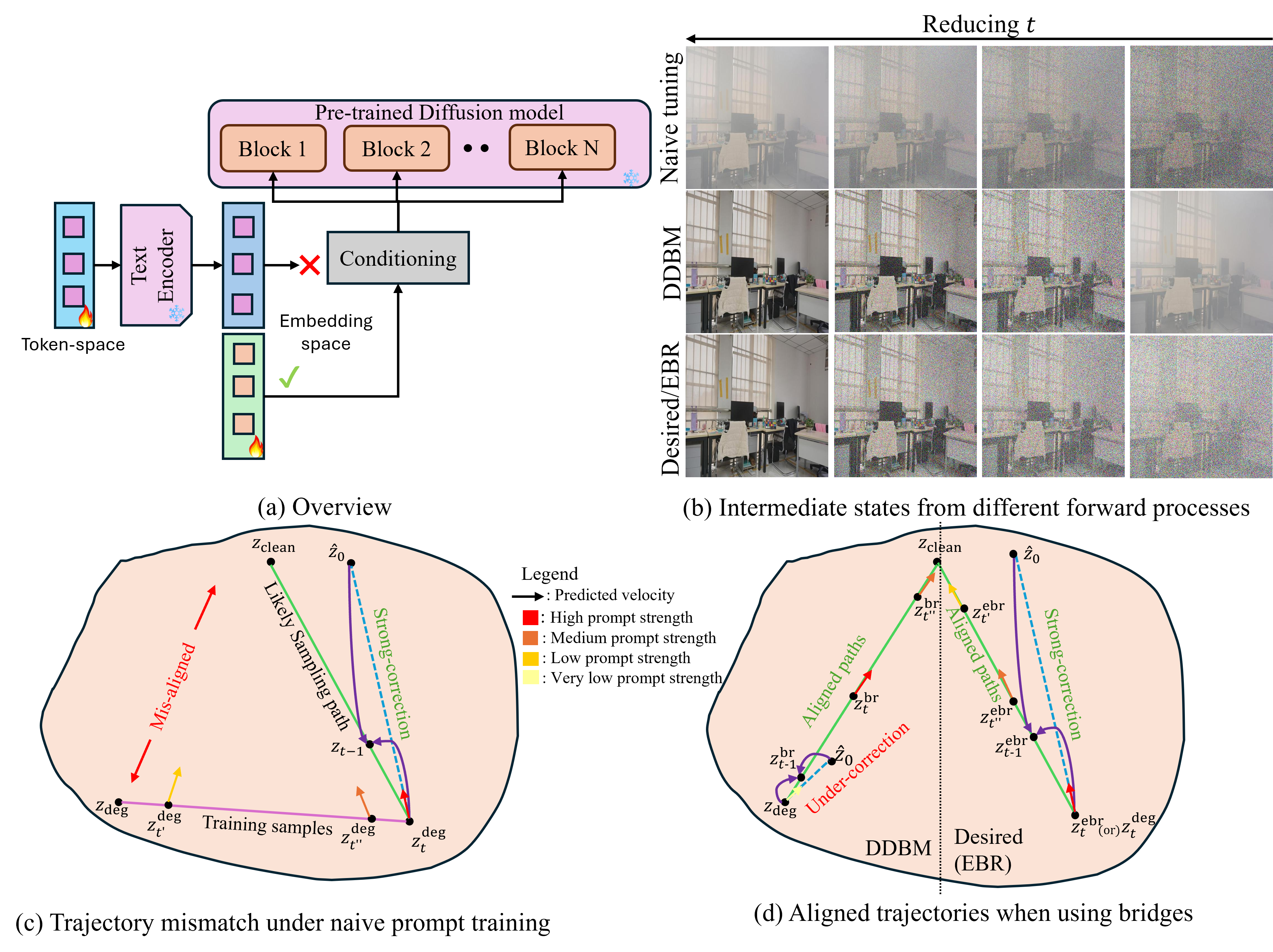}
    \caption{(a) We freeze the diffusion backbone and optimize only the conditioning: token-space prompt optimization fails, while embedding-space (text-encoder output) optimization elicits restoration. (b) Naive tuning yields states anchored at $z_{\text{deg}}$; DDBM~\cite{ddbm} is pinned at both endpoints; our desired/EBR-style~\cite{gcb} bridge starts from noisy degraded inputs and denoises monotonically as the content transitions toward $z_{\text{clean}}$. (c) Naive training sees a different state family than inference, causing trajectory misalignment. (d) Bridge-based training aligns train/test states; DDBM may under-correct early (low noise near $z_{\text{deg}}$), while the desired/EBR bridge enables stronger correction on an aligned path.}

    \label{fig:block}
\end{figure}
\subsection{Guiding pre-trained diffusion models for restoration}
\label{subsec: prompting}

Given paired data $(x_{\text{deg}}, x_{\text{clean}})$, our goal is to produce a restored image/video, $x_{\text{pred}}$, that matches $x_{\text{clean}}$.
Toward this aim, we leverage large pre-trained text-to-image/video diffusion backbones that operate in the latent space of a VAE whose encoder is $\mathcal{E}$.
Let $z_{\text{deg}} = \mathcal{E}(x_{\text{deg}})$ and $z_{\text{clean}} = \mathcal{E}(x_{\text{clean}})$ denote the corresponding VAE latents,
and let $\epsilon\sim\mathcal{N}(0,I)$. Following the flow-matching parameterization used by WAN/FLUX, we construct a noisy degraded input as
\begin{equation}
z_t^{\text{deg}} = \bigl(1-t\bigr)\,z_{\text{deg}} + t\,\epsilon,
\qquad t\in[0,1],
\label{eq:deg_forward_fm}
\end{equation}
Since these diffusion models possess strong priors of the natural image distribution, we ask whether the noisy degraded latent, $z_t^{\text{deg}}$, can be restored by \emph{steering} reverse-time sampling with a restoration instruction. Concretely, we condition the frozen backbone on a text prompt $c$ (e.g., ``Remove the haze from this image and make it clear.'') and perform reverse sampling from timestep $t$. Let $\tau$ denote tokenization and embedding lookup, and let $\mathrm{TE}$ denote the frozen text encoder.
For the prompt $c$, we obtain token embeddings $y=\tau(c) \in\mathbb{R}^{L\times D_\text{in}},$ and context embeddings as
$e = \mathrm{TE}(y)\in\mathbb{R}^{L\times D}$, where $L$ is the number of tokens and $D$ is the embedding dimension. $e$ then enters the frozen backbone through a conditioning pathway (e.g. cross-attention). However, as shown in Fig.~\ref{fig: text_motiv}, such natural-language prompts fail to guide the backbone from intermediate noisy degraded states toward $z_{\text{clean}}$, and instead suppress the added noise, and return to the degraded input. 

We further test whether this limitation is simply due to insufficient expressivity of the text prompt by optimizing the prompt itself (in token-space) while keeping the diffusion backbone and text encoder frozen. Concretely, we form the noisy degraded state $z_t^{\text{deg}}$ as in Eq.~\eqref{eq:deg_forward_fm} and seek prompt tokens that make the frozen backbone predict the clean latent ($z_{\text{clean}}$) from this state. More specifically, we optimize token embeddings $U\in\mathbb{R}^{L\times D_{\text{in}}}$ at the input of the text encoder, producing context embeddings $u=\mathrm{TE}(U) \in R^{L\times D}$ that minimize the backbone's reconstruction error of the clean latent
\begin{equation}
\mathcal{L}_{\text{restore}}
=
\mathbb{E}_{(z_{\text{deg}},z_{\text{clean}}),\,t,\,\epsilon}
\Big[
\bigl\|\hat z_{0,\theta}(z_t^{\text{deg}}, t; u) - z_{\text{clean}}\bigr\|_2^2
\Big],
\label{eq:naive_prompt_loss_x0}
\end{equation}
where $\hat z_{0,\theta} = z_t^{\text{deg}} - tv_\theta(z_t^{\text{deg}},t;u)$ with the expectation over training data pairs. Despite optimizing for $U$, we observe that the loss does not converge and produces a denoised degraded output (Fig.~\ref{fig: text_motiv}).

Surprisingly, bypassing the token space and directly optimizing the conditioning \emph{in context embedding space} elicits a strong restoration response.
Let $p\in\mathbb{R}^{L\times D}$ denote the context embedding vector in the output space of the text encoder fed into the backbone's conditioning pathway.
We now keep the diffusion backbone frozen and optimize $p$ using the same clean sample prediction objective as in Eqn.~\ref{eq:naive_prompt_loss_x0}. Unlike token-space optimization, we find that embedding-space optimization (Fig.~\ref{fig:block}(a)) consistently drives $\hat z_{0,\theta}(z_t^{\text{deg}},t;p)$ toward $z_{\text{clean}}$ and produces visibly restored predictions from the same noisy degraded inputs (Fig.~\ref{fig: text_motiv}).
This suggests that restoration-relevant capability is present in the pre-trained diffusion prior but is not accessible through the standard input token-based interface.

While embedding-space prompt optimization elicits restoration behavior during training, it introduces a train-test \emph{trajectory mismatch} at inference-time sampling.
Specifically, Eq.~\eqref{eq:deg_forward_fm} constructs training states whose signal component remains anchored at $z_{\text{deg}}$, so the learned prompt $p$ is only exposed to the intermediate samples $\{z_t^{\text{deg}}\}_{t\in[0,1]}$ along the ``degraded $\rightarrow$ degraded+noise'' path (see Figs.~\ref{fig:block}(b) and (c)).
During sampling, however, the states at each timestep must progressively move from $z_{\text{deg}}$ toward $z_{\text{clean}}$, and therefore visit intermediate states whose signal component is no longer centered at $z_{\text{deg}}$ (see Fig.~\ref{fig:block}(c)).
This applies $p$ to unseen intermediate states, leading to instability and artifacts (see Sec.~\ref{subsec: ablation}). We address this mismatch in the next section.

\subsection{Addressing Trajectory Mismatch}
\label{subsec: traj_mismatch}

The afore-mentioned problem motivates constructing training trajectories whose intermediate states smoothly transition from degraded to clean, i.e., a \emph{diffusion bridge} between the paired endpoints $z_{\text{deg}}$ and $z_{\text{clean}}$. A bridge is attractive because it provides a sequence of intermediate states that (i) interpolates the signal from $z_{\text{deg}}$ to $z_{\text{clean}}$ and (ii) can be used in both training and inference, thereby mitigating the train-test trajectory mismatch (see Fig.~\ref{fig:block}(d)). Thus, in our setting with a frozen diffusion backbone and a trainable conditioning prompt $p$, the bridge defines compatible training inputs and sampling trajectories, ensuring that $p$ encounters similar intermediate states at training and test time.

Delving deeper, denoising diffusion bridge models (DDBMs)~\cite{ddbm} construct analytically tractable Gaussian distributions of the form
\begin{equation}
q(z_t \mid \zclean,\zdeg)
= \mathcal{N}\!\Big(z_t;\; a_t\,\zdeg + b_t\,\zclean,\; s_t^2 I\Big),
\label{eq:ddbm_bridge}
\end{equation}
with $(a_t,b_t,s_t)$ defined by the chosen bridge schedule~\cite{ddbm} at time $t$. This yields a closed-form expression for sampling the input from the forward process as
\begin{equation}
z^{\text{br}}_t = a_t\,\zdeg + b_t\,\zclean + s_t\,\epsilon,
\qquad \epsilon\sim\mathcal{N}(0,I).
\label{eq:ddbm_sample}
\end{equation}
However, WAN/FLUX models expect an input of the form in Eqn.~\ref{eq:deg_forward_fm}. Additionally, the noise level of the bridge needs to be matched with that of the frozen backbone. To achieve this, we first view $a_t\,\zdeg + b_t\,\zclean$ as the signal component of the input and $\epsilon$ as the sampled noise. If we multiply by $(1-\sigma_t)$, where $\sigma_t$ is the desired noise coefficient for WAN/FLUX, we get
\begin{equation}
y^{\text{br}}_t = (1-\sigma_t)(a_t\,\zdeg + b_t\,\zclean) + (1-\sigma_t)s_t\,\epsilon.
\label{eq:ddbm_reparam1}
\end{equation}
Now, defining $\sigma_t=\frac{s_t}{1+s_t}$ implies $(1-\sigma_t)s_t=\sigma_t$ and yields
\begin{equation}
y^{\text{br}}_t = (1-\sigma_t)(a_t\,\zdeg + b_t\,\zclean) + \sigma_t\,\epsilon.
\label{eq:ddbm_reparam2}
\end{equation}
which is now in the form of Eqn.~\ref{eq:deg_forward_fm}. We now optimize the prompt so that the frozen backbone, when conditioned on $p$, predicts the clean endpoint from $y^{\text{br}}_t$:
\begin{equation}
\min_{p}\;
\mathbb{E}_{(z_{\text{deg}},z_{\text{clean}}),\,t,\,\epsilon}
\Big[
\bigl\|\hat z_{0,\theta}(y^{\text{br}}_t,t;p) - z_{\text{clean}}\bigr\|_2^2
\Big],
\label{eq:prompt_bridge_obj}
\end{equation}
where $\hat z_{0,\theta}(\cdot)$ is the clean estimate predicted by the frozen backbone (Sec.~\ref{subsec: prompting}). Intuitively, the above process can be viewed as the prompt observing various intermediate signal components of the form $a_t\,\zdeg + b_t\,\zclean$ and providing the necessary guidance to restore them. During sampling, $\hat z_{0,\theta}(\cdot)$ is used to step through the reverse trajectory~\cite{ddbm}, ensuring the prompt encounters similar intermediate samples as in training (see Fig.~\ref{fig:block}(d)).

Despite the alignment benefit, using the above DDBM-like bridge for training our prompts with frozen backbones leads to two key problems. First, the bridge variance $s_t^2$ is typically \emph{small near both endpoints} and peaks in the middle (see row $2$ of Fig.~\ref{fig:block}(b).
This means, near the degraded endpoint the state is almost deterministic and close to $\zdeg$, leaving limited stochasticity/uncertainty for the frozen generative prior to ``re-route'' the trajectory early (see Fig.~\ref{fig:block}(d)). Empirically, this significantly weakens the prompt's restoration capability at the beginning of sampling and leads to under-correction of degradations (see Sec.~\ref{subsec: ablation}). Second, deterministic integration (ODE-based solvers) from a fixed endpoint can yield overly smooth trajectories~\cite{ddbm,dbim}. DDBM mitigates this with stochastic ``churn''~\cite{ddbm}, but churn increases neural function evaluations (NFEs), which is expensive for large backbones (WAN/FLUX) and can still yield suboptimal performance~\cite{dbim}. Thus, we require a bridge-like framework whose \emph{starting state already contains noise on top of the degraded input} and whose noise level \emph{decreases monotonically} as the signal transitions toward clean (see row $3$ of Fig.~\ref{fig:block}(b)). 

Concretely, we want intermediate states of the form
\begin{equation}
z^{\text{br}}_t
=
\mu(t) + s(t)\,\epsilon,
\qquad
\mu(t) = \lambda(t)\,z_{\text{deg}} + \bigl(1-\lambda(t)\bigr)\,z_{\text{clean}},
\qquad
\epsilon\sim\mathcal{N}(0,I),
\label{eq:desired_bridge}
\end{equation}
where $t\in[0,T_0]$ (with $T_0\le 1$), $\lambda(0)=0$, $\lambda(T_0)=1$, and $s(t)$ \emph{increases} with $t$.
This construction (i) exposes the prompt early on to noisy degraded states allowing strong restoration, while (ii) gradually moving towards cleaner states at later times, avoiding the train-test trajectory mismatch (see Fig.~\ref{fig:block}(d)). For this purpose, we adopt the energy-oriented diffusion bridge (EBR)~\cite{gcb} which constructs bridge states as
\begin{equation}
z^{\text{ebr}}_t
=
(1-t)\Bigl[\Bigl(1-\frac{t}{T_0}\Bigr)\,z_{\text{clean}} + \Bigl(\frac{t}{T_0}\Bigr)\,z_{\text{deg}}\Bigr]
+ t\,\epsilon,
\label{eq:gcb_final}
\end{equation}
which satisfies our desired endpoint behaviour. Importantly, our contribution is \emph{not} the bridge schedule itself, but showing that this monotone noisy-to-clean bridge is the missing ingredient that makes our prompt-only restoration (Sec.~\ref{subsec: prompting}) feasible on frozen large diffusion models.

Thus, we leverage the EBR forward construction to sample training states $z^{\text{ebr}}_t$ and train $p$ with the clean endpoint prediction objective 
\begin{equation}
\min_{p}\;
\mathbb{E}_{(z_{\text{deg}},z_{\text{clean}}),\,t,\,\epsilon}
\Big[
\bigl\|\hat z_{0,\theta}(z^{\text{ebr}}_t,t;p) - z_{\text{clean}}\bigr\|_2^2
\Big],
\label{eq:prompt_desired_obj}
\end{equation}
where $\hat z_{0,\theta}(\cdot)$ is the clean estimate using $p$ with the frozen backbone. At inference, we run a deterministic sampler (DDIM-like~\cite{ddim}) using the same bridge schedule so that the intermediate states visited during sampling match those seen during training, mitigating trajectory mismatch and enabling successful prompt training (see Sec.~\ref{subsec: ablation}).

\subsection{Design choices}
\label{subsec: design_choices}

\noindent\textbf{One prompt per degradation.}
We learn a separate conditioning prompt $p^{(k)}$ for each degradation type $k$ (e.g., haze, rain, low-light), while keeping the diffusion backbone frozen.
This still follows the AiOR setup where a single backbone is reused across degradations and task-specific information is provided as conditioning. Here, the conditioning is learned prompts while in AiOR approaches such as  PixWizard~\cite{pixwizard} and DFPIR~\cite{dfpir}, degradation-specific text is used to guide a single backbone.
For addressing mixed degradations, we combine prompts by averaging their predicted velocities during sampling, enabling joint restoration without additional training.

\noindent \textbf{Prompt injection and residual learning.}
Rather than optimizing the text-encoder output directly, we learn a \emph{residual} prompt on top of the null-text context and inject it at the conditioning interface (e.g. input to cross-attention). We found this design is more parameter-efficient and improves performance compared to directly optimizing the text-encoder output (see Sec.~\ref{subsec: ablation}). We provide the specific-architectural details in the supplementary.

\noindent\textbf{Choosing the start noise level $T_0$.}
We set the maximum noising level $T_0$ used in training (Eq.~\ref{eq:gcb_final}) based on degradation severity.
Milder degradations typically require lower noise to restore, whereas severe degradations benefit from a larger $T_0$ to provide the frozen prior sufficient flexibility to correct the input.
In our training data, low-light exhibits the most severe corruption levels. We therefore tune $T_0$ on low-light and find $T_0=0.4$ performs best (Sec.~\ref{subsec: ablation}).
For simplicity and robustness, we use this same $T_0$ for all degradations, ensuring adequate headroom for difficult cases.

\renewcommand{\best}[1]{#1}
\renewcommand{\second}[1]{#1}

\begin{table}[t]
\centering
\vspace{-5pt}
\caption{Comparisons of mean within-distribution performance of our approach on the WAN and FLUX models with state-of-the-art image and video AiOR approaches.}
\scriptsize
\setlength{\tabcolsep}{3.2pt}
\renewcommand{\arraystretch}{1.05}

\adjustbox{max width=\textwidth}{%
\begin{tabular}{l|cccccc|ccccccc}
\toprule
\multicolumn{1}{c|}{\multirow{2}{*}{\textbf{Methods}}} &
\multicolumn{6}{c|}{\textbf{Image restoration (mean)}} &
\multicolumn{7}{c}{\textbf{Video restoration (mean)}} \\
\cmidrule(lr){2-7}\cmidrule(lr){8-14}
& \textbf{P$\uparrow$} & \textbf{S$\uparrow$} & \textbf{L$\downarrow$} & \textbf{D$\downarrow$} & \textbf{C$\uparrow$} & \textbf{M$\uparrow$}
& \textbf{P$\uparrow$} & \textbf{S$\uparrow$} & \textbf{L$\downarrow$} & \textbf{D$\downarrow$} & \textbf{C$\uparrow$} & \textbf{M$\uparrow$} & \textbf{Do$\uparrow$} \\
\midrule
DCPT      & 24.82 & 0.812 & 0.178 & 0.119 & 0.542 & 61.57  & 19.50&0.715&0.285&0.167&0.462&54.75&0.420 \\
DFPIR     & 24.96 & 0.813 & 0.177 & 0.117 & 0.557 & 61.87  & 20.16&0.743&0.247&0.151&0.493&58.64&0.462 \\
AutoDIR   & 26.29 & 0.854 & 0.138 & 0.094 & 0.591 & 64.01  & 19.57&0.739&0.236&0.144&0.509&60.17&0.490 \\
PixWizard & 19.40 & 0.608 & 0.229 & 0.133 & 0.627 & 64.30  & 17.73&0.612&0.248&0.137&0.581&63.26&0.521 \\
FoundIR   & 18.36 & 0.672 & 0.281 & 0.193 & 0.497 & 55.58  & 17.30&0.672&0.309&0.223&0.434&51.99&0.419 \\
FLUX (Ours)    & 23.30 & 0.750 & 0.173 & 0.119 & 0.552 & 64.11  & --    & --    & --    & --    & --    & --    & --    \\
\midrule
ViWS-Net  & --    & --    & --    & --    & --    & --     & 19.76&0.716&0.310&0.241&0.381&45.87&0.375 \\
AverNet   & --    & --    & --    & --    & --    & --     & 21.75&0.799&0.250&0.169&0.357&38.88&0.349 \\
WAN (Ours)& --    & --    & --    & --    & --    & --     & 20.89&0.701&0.184&0.118&0.534&64.19&0.571 \\
\bottomrule
\end{tabular}%
}
\vspace{-10pt}
\label{tab: quant_wd}
\end{table}

\begin{table}[t]
\centering
\caption{Quantitative comparisons of our prompt learning approach on the FLUX model with state-of-the-art AiOR approaches for images from OOD, mixed and unseen degradations. H-Haze, R-Rain, S-Snow, B-Blur, L-Low-light, and U-Unseen.}
\vspace{-5pt}
\scriptsize

\adjustbox{max width=\textwidth}{%
\begin{tabular}{l|cccccc|cccccc|cccccc}
\toprule
\multicolumn{1}{c|}{\multirow{2}{*}{\textbf{Methods}}} &
\multicolumn{6}{c|}{\textbf{HazeRD (H)}} &
\multicolumn{6}{c|}{\textbf{LHP (R)}} &
\multicolumn{6}{c}{\textbf{WeatherBench (S)}} \\
\cmidrule(lr){2-7}\cmidrule(lr){8-13}\cmidrule(lr){14-19}
& \textbf{P$\uparrow$} & \textbf{S$\uparrow$} & \textbf{L$\downarrow$} & \textbf{D$\downarrow$} & \textbf{C$\uparrow$} & \textbf{M$\uparrow$}
& \textbf{P$\uparrow$} & \textbf{S$\uparrow$} & \textbf{L$\downarrow$} & \textbf{D$\downarrow$} & \textbf{C$\uparrow$} & \textbf{M$\uparrow$}
& \textbf{P$\uparrow$} & \textbf{S$\uparrow$} & \textbf{L$\downarrow$} & \textbf{D$\downarrow$} & \textbf{C$\uparrow$} & \textbf{M$\uparrow$} \\
\midrule
DCPT      & 13.98 & 0.774 & 0.214 & 0.156 & 0.620 & 66.16
          & 27.49 & 0.828 & 0.196 & 0.135 & 0.480 & 58.87
          & 21.31 & \second{0.753} & 0.244 & 0.173 & 0.363 & \best{49.96} \\
DFPIR     & 15.73 & 0.795 & \second{0.181} & \second{0.123} & 0.681 & 68.24
          & \second{29.96} & \second{0.854} & 0.177 & 0.124 & 0.475 & 57.21
          & 17.77 & 0.693 & 0.272 & 0.184 & 0.332 & 44.12 \\
AutoDIR   & \second{16.59} & \best{0.843} & \best{0.164} & \best{0.114} & 0.679 & \best{69.38}
          & 28.60 & 0.840 & 0.183 & 0.125 & 0.537 & \second{59.00}
          & 21.14 & 0.746 & 0.245 & 0.169 & 0.386 & \second{47.90} \\
PixWizard & 15.16 & 0.638 & 0.249 & 0.159 & \best{0.696} & 68.96
          & 20.88 & 0.643 & 0.230 & 0.153 & \best{0.615} & \best{59.30}
          & \best{22.40} & 0.691 & \best{0.197} & \best{0.128} & \best{0.447} & 45.43 \\
FoundIR   & 14.41 & \second{0.810} & 0.236 & 0.182 & 0.611 & 68.18
          & \best{30.12} & \best{0.868} & \best{0.155} & \best{0.107} & 0.527 & 56.43
          & 21.60 & \best{0.769} & \second{0.240} & \second{0.165} & \second{0.410} & 45.23 \\
FLUX (Ours)    & 16.70 & 0.705 & 0.233 & 0.162 & \second{0.685} & \second{69.01}
          & 26.44 & 0.816 & 0.163 & 0.108 & 0.542 & 58.09
          & 22.07 & 0.738 & 0.264 & 0.178 & 0.389 & 46.52 \\
\bottomrule
\end{tabular}%
}

\adjustbox{max width=\textwidth}{%
\begin{tabular}{l|cccccc|cccccc|cccccc}
\toprule
\multicolumn{1}{c|}{\multirow{2}{*}{\textbf{Methods}}} &
\multicolumn{6}{c|}{\textbf{4KRD (B)}} &
\multicolumn{6}{c|}{\textbf{SICE (L)}} &
\multicolumn{6}{c}{\textbf{TOLED (U)}} \\
\cmidrule(lr){2-7}\cmidrule(lr){8-13}\cmidrule(lr){14-19}
& \textbf{P$\uparrow$} & \textbf{S$\uparrow$} & \textbf{L$\downarrow$} & \textbf{D$\downarrow$} & \textbf{C$\uparrow$} & \textbf{M$\uparrow$}
& \textbf{P$\uparrow$} & \textbf{S$\uparrow$} & \textbf{L$\downarrow$} & \textbf{D$\downarrow$} & \textbf{C$\uparrow$} & \textbf{M$\uparrow$}
& \textbf{P$\uparrow$} & \textbf{S$\uparrow$} & \textbf{L$\downarrow$} & \textbf{D$\downarrow$} & \textbf{C$\uparrow$} & \textbf{M$\uparrow$} \\
\midrule
DCPT      & 23.48 & 0.828 & 0.193 & 0.142 & 0.509 & 58.33
          & 15.29 & \second{0.671} & \best{0.228} & \best{0.144} & \best{0.618} & \second{66.30}
          & 15.14 & 0.630 & 0.323 & 0.217 & 0.419 & 46.63 \\
DFPIR     & \second{27.72} & \best{0.870} & 0.150 & 0.119 & 0.527 & 61.83
          & \best{16.41} & \best{0.722} & \second{0.256} & \second{0.165} & 0.578 & \best{68.40}
          & 11.92 & 0.558 & 0.337 & 0.224 & 0.429 & 45.71 \\
AutoDIR   & 27.46 & \second{0.865} & \second{0.143} & \second{0.112} & 0.563 & 65.26
          & 13.55 & 0.620 & 0.284 & 0.183 & 0.602 & 65.91
          & 16.99 & 0.669 & 0.270 & 0.192 & 0.450 & 49.71 \\
PixWizard & 19.86 & 0.584 & 0.193 & 0.126 & \second{0.652} & \second{69.10}
          & 12.33 & 0.432 & 0.373 & 0.217 & \second{0.613} & 63.79
          & \second{24.41} & \second{0.731} & \second{0.196} & \best{0.151} & \best{0.549} & \best{61.30} \\
FoundIR   & \best{27.87} & 0.865 & 0.150 & 0.124 & 0.528 & 60.81
          & 10.93 & 0.449 & 0.395 & 0.253 & 0.557 & 60.50
          & \best{28.08} & \best{0.824} & \best{0.191} & \second{0.165} & 0.429 & 48.64 \\
FLUX (Ours)    & 22.80 & 0.702 & 0.137 & 0.094 & 0.674 & 72.67
          & 15.39 & 0.566 & 0.328 & 0.208 & 0.585 & 66.08
          & 18.05 & 0.671 & 0.229 & 0.173 & 0.515 & 58.05 \\
\bottomrule
\end{tabular}%
}

\adjustbox{max width=\textwidth}{%
\begin{tabular}{l|cccccc|cccccc|cc|cccccc}
\toprule
\multicolumn{1}{c|}{\multirow{2}{*}{\textbf{Methods}}} &
\multicolumn{6}{c|}{\textbf{POLED (U)}} &
\multicolumn{6}{c|}{\textbf{CDD (S+H)}} &
\multicolumn{2}{c|}{\textbf{LOLBlur (L+B)}} &
\multicolumn{6}{c}{\textbf{Average}} \\
\cmidrule(lr){2-7}\cmidrule(lr){8-13}\cmidrule(lr){14-15}\cmidrule(lr){16-21}
& \textbf{P$\uparrow$} & \textbf{S$\uparrow$} & \textbf{L$\downarrow$} & \textbf{D$\downarrow$} & \textbf{C$\uparrow$} & \textbf{M$\uparrow$}
& \textbf{P$\uparrow$} & \textbf{S$\uparrow$} & \textbf{L$\downarrow$} & \textbf{D$\downarrow$} & \textbf{C$\uparrow$} & \textbf{M$\uparrow$}
& \textbf{C$\uparrow$} & \textbf{M$\uparrow$}
& \textbf{P$\uparrow$} & \textbf{S$\uparrow$} & \textbf{L$\downarrow$} & \textbf{D$\downarrow$} & \textbf{C$\uparrow$} & \textbf{M$\uparrow$} \\
\midrule
DCPT      & 9.06 & 0.364 & 0.613 & 0.462 & 0.340 & 42.93
          & 13.85 & 0.713 & 0.308 & 0.223 & 0.553 & 61.17
          & 0.374 & 39.58
          & 17.45 & 0.695 & 0.289 & 0.206 & 0.475 & 54.43 \\
DFPIR     & 8.93 & 0.351 & 0.657 & 0.461 & 0.324 & 41.82
          & 14.95 & 0.721 & 0.305 & 0.214 & 0.555 & 60.85
          & 0.344 & 39.13
          & 17.92 & 0.695 & 0.291 & 0.201 & 0.471 & 54.14 \\
AutoDIR   & 11.69 & 0.435 & \second{0.564} & \second{0.417} & \second{0.348} & \best{47.44}
          & \best{19.76} & \best{0.827} & \best{0.223} & \best{0.145} & \second{0.668} & \second{66.35}
          & 0.404 & 43.14
          & 19.47 & 0.730 & 0.259 & 0.182 & 0.515 & 57.12 \\
PixWizard & \second{17.62} & 0.462 & \best{0.477} & \best{0.341} & \best{0.387} & \second{43.68}
          & 16.19 & 0.509 & 0.250 & 0.162 & \best{0.675} & 66.27
          & \best{0.536} & \best{56.11}
          & 18.60 & 0.586 & 0.270 & 0.179 & 0.574 & 59.32 \\
FoundIR   & 16.32 & \second{0.478} & 0.658 & 0.531 & 0.280 & 27.94
          & \second{18.07} & \second{0.805} & \second{0.228} & \second{0.154} & 0.620 & 63.09
          & 0.335 & 35.38
          & 20.92 & 0.733 & 0.281 & 0.210 & 0.477 & 51.80 \\
FLUX (Ours)    & 18.52 & 0.474 & 0.647 & 0.442 & 0.300 & 38.54
          & 16.66 & 0.591 & 0.253 & 0.171 & 0.651 & 67.34
          & 0.425 & 48.21
          & 19.57 & 0.657 & 0.281 & 0.192 & 0.529 & 58.27 \\
\bottomrule
\end{tabular}%
}
\vspace{-10pt}
\label{tab:quant_img_ood}
\end{table}

\begin{table}[t]
\centering
\caption{Quantitative comparisons of our prompt learning approach on the WAN model with state-of-the-art image and video restoration approaches for the task of all-in-one video restoration on OOD datasets. H-Haze, R-Rain, S-Snow, B-Blur, and L-Low-light}
\vspace{-5pt}
\scriptsize
\setlength{\tabcolsep}{2.7pt}
\newcommand{\rowh}{\rule{0pt}{2.15ex}}

\adjustbox{max width=\textwidth}{%
\begin{tabular}{l|ccc|ccccccc|ccc|ccc}
\toprule
\multicolumn{1}{c|}{\multirow{2}{*}{\textbf{Methods}}} &
\multicolumn{3}{c|}{\textbf{RHVD (H)}} &
\multicolumn{7}{c|}{\textbf{LasVR (R)}} &
\multicolumn{3}{c|}{\textbf{NTURain (R)}} &
\multicolumn{3}{c}{\textbf{AAURainSnow (S)}} \\
\cmidrule(lr){2-4}\cmidrule(lr){5-11}\cmidrule(lr){12-14}\cmidrule(lr){15-17}
& \textbf{$\uparrow$C} & \textbf{M$\uparrow$} & \textbf{Do$\uparrow$}
& \textbf{P$\uparrow$} & \textbf{S$\uparrow$} & \textbf{L$\downarrow$} & \textbf{D$\downarrow$} & \textbf{C$\uparrow$} & \textbf{M$\uparrow$} & \textbf{Do$\uparrow$}
& \textbf{C$\uparrow$} & \textbf{M$\uparrow$} & \textbf{Do$\uparrow$}
& \textbf{C$\uparrow$} & \textbf{M$\uparrow$} & \textbf{Do$\uparrow$} \\
\midrule
DCPT      \rowh & 0.558&59.66&0.530  & 29.16&0.846&0.176&0.141&0.544&63.04&0.628  & 0.599&63.37&0.571  & 0.432&47.60&0.321 \\
DFPIR     \rowh & 0.564&59.67&0.543  & 31.94&0.909&0.101&0.097&0.546&61.94&0.632  & 0.594&63.39&0.545  & 0.439&47.59&0.340 \\
AutoDIR   \rowh & 0.579&61.25&0.522  & 29.39&0.818&0.195&0.142&0.570&64.12&0.616  & 0.627&65.34&0.563  & 0.459&49.51&0.301 \\
PixWizard \rowh & 0.595&60.10&0.571  & 23.20&0.655&0.168&0.121&0.625&61.84&0.641  & 0.643&64.38&0.548  & 0.459&45.31&0.299 \\
FoundIR   \rowh & 0.558&60.75&0.581  & 31.54&0.883&0.148&0.123&0.530&61.05&0.655  & 0.600&61.95&0.573  & 0.446&45.86&0.319 \\
\midrule
ViWS-Net  \rowh & 0.529&58.72&0.554  & 29.79&0.847&0.135&0.118&0.527&61.50&0.657  & 0.513&58.21&0.594  & 0.420&48.13&0.359 \\
AverNet   \rowh & 0.488&50.62&0.552  & 30.80&0.863&0.176&0.149&0.472&50.43&0.582  & 0.533&52.98&0.545  & 0.416&43.62&0.335 \\
WAN (Ours)\rowh & 0.598&63.85&0.653  & 27.95&0.775&0.153&0.105&0.564&62.03&0.664  & 0.583&62.75&0.615  & 0.463&49.42&0.373 \\
\bottomrule
\end{tabular}%
}

\adjustbox{max width=\textwidth}{%
\begin{tabular}{l|ccccccc|ccc|ccccccc}
\toprule
\multicolumn{1}{c|}{\multirow{2}{*}{\textbf{Methods}}} &
\multicolumn{7}{c|}{\textbf{4KRD (B)}} &
\multicolumn{3}{c|}{\textbf{Lol-iPhone (L)}} &
\multicolumn{7}{c}{\textbf{Average}} \\
\cmidrule(lr){2-8}\cmidrule(lr){9-11}\cmidrule(lr){12-18}
& \textbf{P$\uparrow$} & \textbf{S$\uparrow$} & \textbf{L$\downarrow$} & \textbf{D$\downarrow$} & \textbf{C$\uparrow$} & \textbf{M$\uparrow$} & \textbf{Do$\uparrow$}
& \textbf{C$\uparrow$} & \textbf{M$\uparrow$} & \textbf{Do$\uparrow$}
& \textbf{P$\uparrow$} & \textbf{S$\uparrow$} & \textbf{L$\downarrow$} & \textbf{D$\downarrow$} & \textbf{C$\uparrow$} & \textbf{M$\uparrow$} & \textbf{Do$\uparrow$} \\
\midrule
DCPT      \rowh & 23.16&0.819&0.202&0.145&0.504&57.72&0.500  & 0.354&41.34&0.335  & 26.16&0.833&0.189&0.143&0.498&55.46&0.481 \\
DFPIR     \rowh & 27.19&0.863&0.154&0.120&0.522&61.36&0.544  & 0.355&44.87&0.385  & 29.57&0.886&0.128&0.108&0.503&56.47&0.498 \\
AutoDIR   \rowh & 26.94&0.858&0.144&0.110&0.564&65.86&0.618  & 0.374&44.18&0.372  & 28.16&0.838&0.169&0.126&0.529&58.38&0.499 \\
PixWizard \rowh & 19.77&0.587&0.196&0.124&0.647&68.60&0.612  & 0.441&48.00&0.373  & 21.48&0.621&0.182&0.122&0.568&58.04&0.507 \\
FoundIR   \rowh & 27.36&0.856&0.155&0.124&0.524&60.28&0.588  & 0.370&40.53&0.397  & 29.45&0.869&0.151&0.123&0.505&55.07&0.519 \\
\midrule
ViWS-Net  \rowh & 25.11&0.801&0.195&0.150&0.523&56.79&0.519  & 0.333&39.52&0.360  & 27.45&0.824&0.165&0.134&0.474&53.81&0.507 \\
AverNet   \rowh & 28.28&0.868&0.149&0.113&0.470&50.81&0.586  & 0.304&35.12&0.423  & 29.54&0.865&0.162&0.131&0.447&47.26&0.504 \\
WAN (Ours)\rowh & 20.98&0.625&0.172&0.117&0.608&69.44&0.648  & 0.328&41.67&0.387  & 24.46&0.700&0.162&0.111&0.524&58.19&0.557 \\
\bottomrule
\end{tabular}%
}
\vspace{-15pt}
\label{tab:quant_video_ood}
\end{table}

\section{Experiments}
\label{sec: experiments}

In this section, we discuss key implementation details, experimental setup, comparisons with other methods, and ablations on our proposed prompts.

\subsection{Implementation Details}
\label{subsec: impl}

We conduct our prompt-learning approach on WAN~\cite{wan} text-to-video model comprising $1.3$ billion parameters and FLUX1 Dev~\cite{flux} text-to-image model consisting of $12$ billion parameters. For WAN, we learn $226$ prompt tokens of dimension $1536$, and for FLUX we learn $512$ prompt tokens of dimension $3072$. We optimize prompts using AdamW~\cite{adamw} with learning rate $5\times10^{-4}$ and batch size $2$, training for $20$k (WAN) and $40$k (FLUX) iterations. All experiments are run on $2\times$ NVIDIA H100 GPUs (80GB).

\subsection{Setup}
\label{subsec: setup}

We train one prompt per degradation on a single in-domain (ID) dataset, and report performance on (i) ID test sets, (ii) out-of-distribution (OOD) datasets, and (iii) mixed and unseen degradations. This evaluation protocol is similar to those followed in~\cite{gendeg, foundir}.

\noindent\textbf{Image restoration (FLUX).} We train prompts for dehazing, desnowing, deraining, low-light enhancement, and deblurring using:
RESIDE~\cite{reside} ($72135$ train images, haze),
Snow100K~\cite{snow100k} ($50000$, snow),
Rain13K~\cite{mprnet} ($13711$, rain),
LOLv1~\cite{lolv1} ($485$, low-light),
and GoPro~\cite{gopro} ($2103$, blur).
For OOD testing, we evaluate on HazeRD~\cite{hazerd} (haze),
LHP~\cite{lhprain} (rain),
WeatherBench~\cite{weatherbench} (snow),
4KRD~\cite{4krd} (blur),
and SICE~\cite{sice} (enhancement).
To assess performance on mixed and unseen degradations, we use TOLED and POLED~\cite{toledpoled} for under-display camera restoration (unseen), haze+snow split of CDD~\cite{cdd}, and LOLBlur~\cite{lolblur} for low-light+blur.

\noindent\textbf{Video restoration (WAN).} We train prompts for dehazing, deraining, desnowing, low-light enhancement, and deblurring using:
REVIDE~\cite{revide} ($40$ train clips, haze),
NTU-Rain (synthetic)~\cite{spac} ($25$, rain),
RVSD (synthetic)~\cite{rsvd} ($77$, snow),
SDSD~\cite{sdsd} ($58$, low-light),
and GoPro~\cite{gopro} ($22$, blur).
For OOD testing, we evaluate on RHVD~\cite{rhvd} (haze),
NTU-Rain (real)~\cite{spac} and LasVR~\cite{lasvr} (rain), AAURainSnow~\cite{aau} (snow),
Lol-iPhone~\cite{loliphone} (low-light),
and 4KRD~\cite{4krd} (blur). During test time, we evaluate on the first $33$ frames of the test videos due to compute constraints.

Among the above, LHP, WeatherBench, TOLED, POLED, LOLBlur, REVIDE, NTU-Rain (real), SDSD, RHVD, AAURainSnow and Lol-iPhone contain real world degradations. We provide the full per-dataset sample counts and details in the supplementary.

\begin{figure*}[t]
    \centering
    \small
    \setlength{\tabcolsep}{1pt}
    \begin{tabular}{cccccccc}
    &Input&DFPIR&AutoDIR&PixWizard&FoundIR&FLUX (Ours)\\
          \rotatebox[origin=c]{90}{TOLED\hspace{-32pt}}&\includegraphics[height=1.37cm, width=1.9cm]{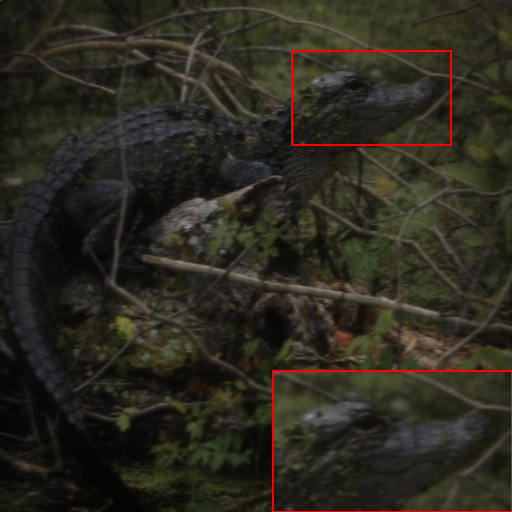}&\includegraphics[height=1.37cm, width=1.9cm]{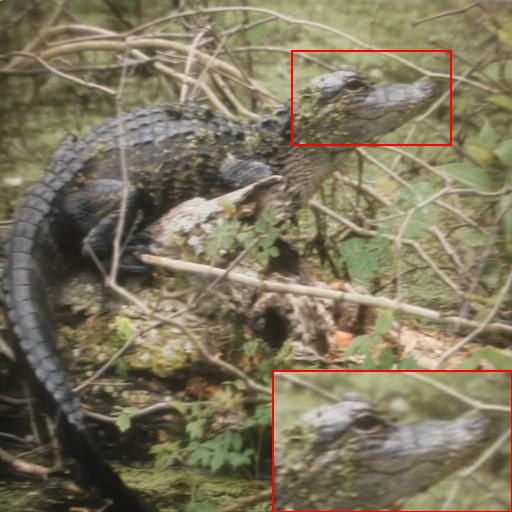}&\includegraphics[height=1.37cm, width=1.9cm]{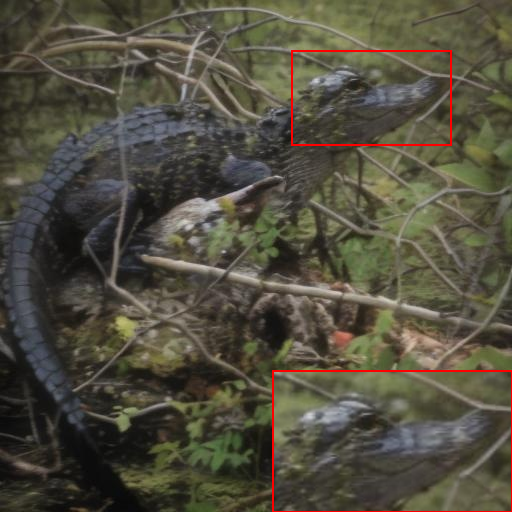}&\includegraphics[height=1.37cm, width=1.9cm]{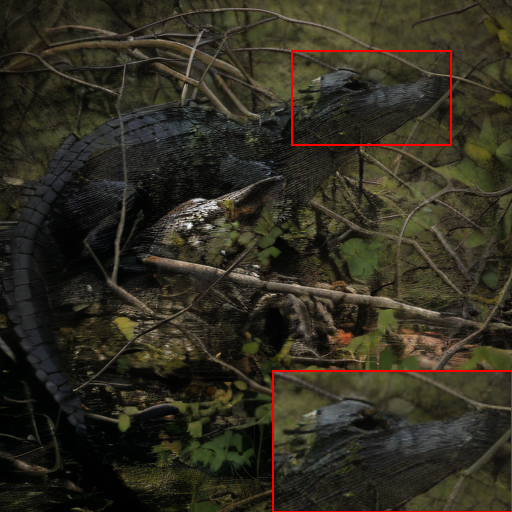}&\includegraphics[height=1.37cm, width=1.9cm]{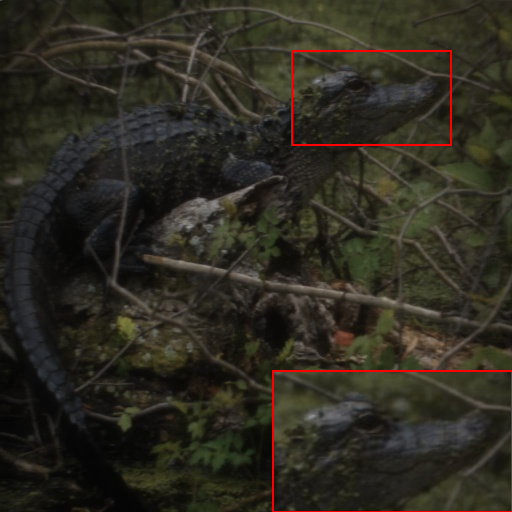}&\includegraphics[height=1.37cm, width=1.9cm]{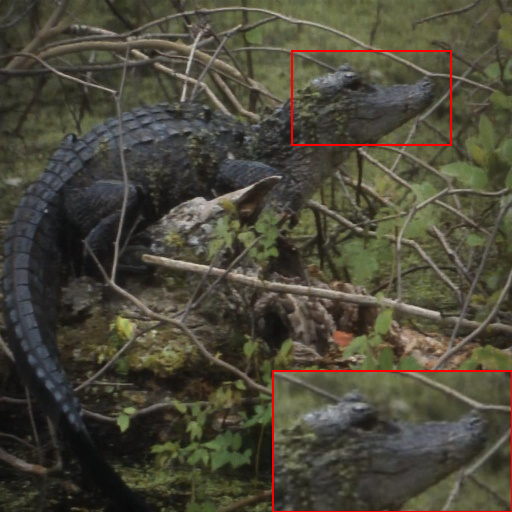}  \\

         
         \rotatebox[origin=c]{90}{LHP\hspace{-32pt}}&\includegraphics[height=1.37cm, width=1.9cm]{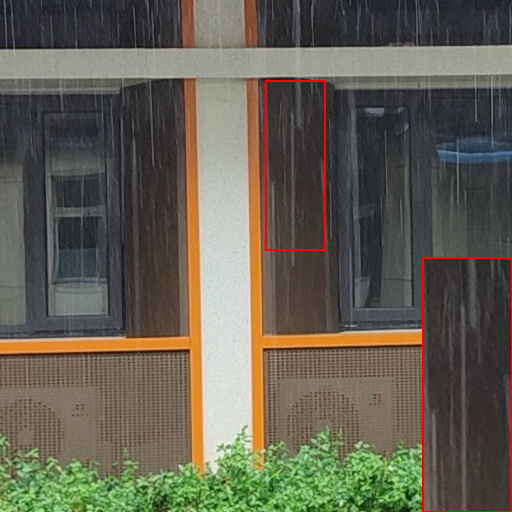}&\includegraphics[height=1.37cm, width=1.9cm]{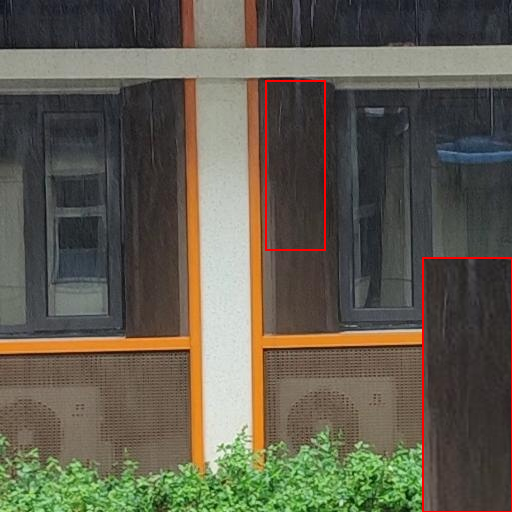}&\includegraphics[height=1.37cm, width=1.9cm]{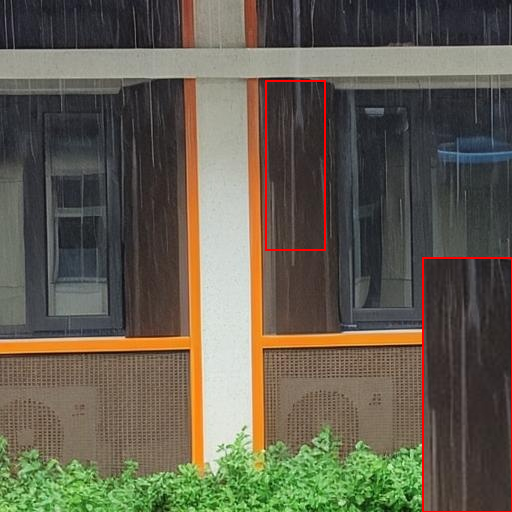}&\includegraphics[height=1.37cm, width=1.9cm]{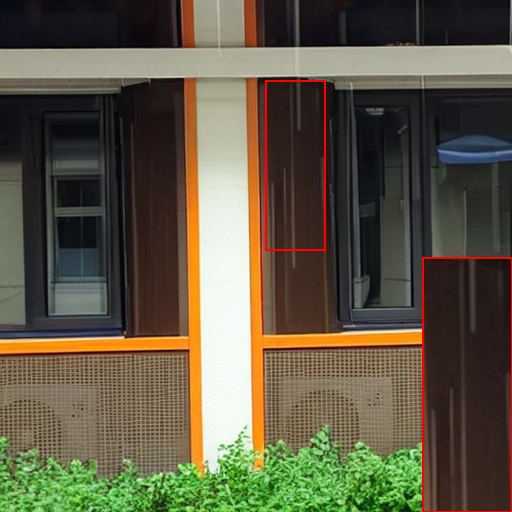}&\includegraphics[height=1.37cm, width=1.9cm]{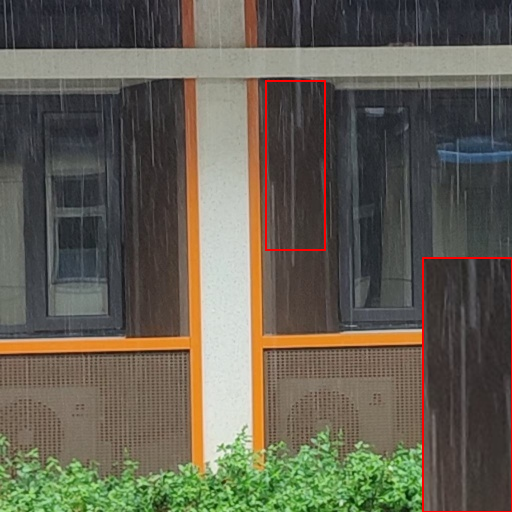}&\includegraphics[height=1.37cm, width=1.9cm]{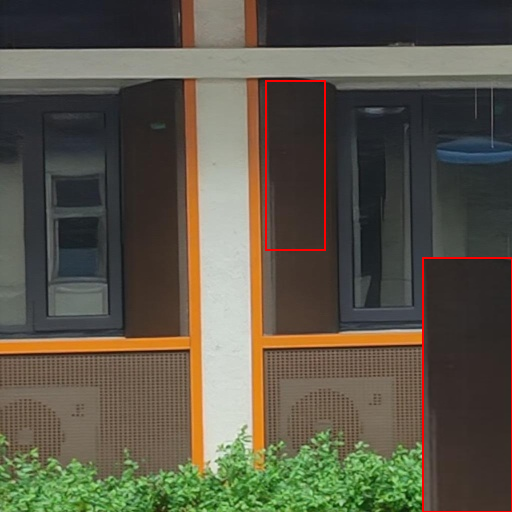}  \\

         \rotatebox[origin=c]{90}{LOLBlur\hspace{-32pt}}&\includegraphics[height=1.37cm, width=1.9cm]{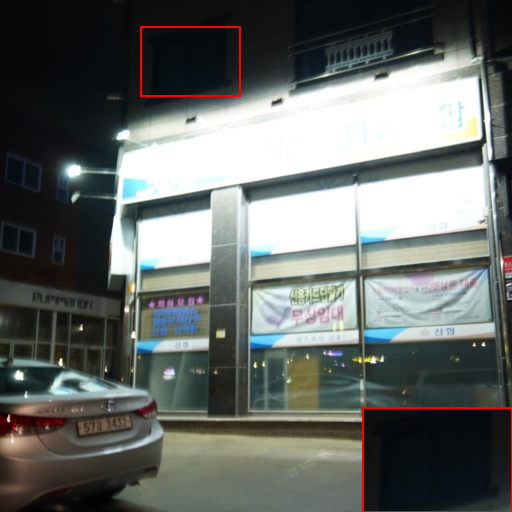}&\includegraphics[height=1.37cm, width=1.9cm]{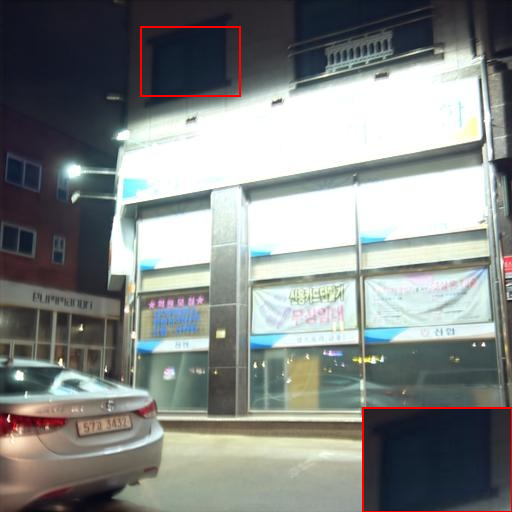}&\includegraphics[height=1.37cm, width=1.9cm]{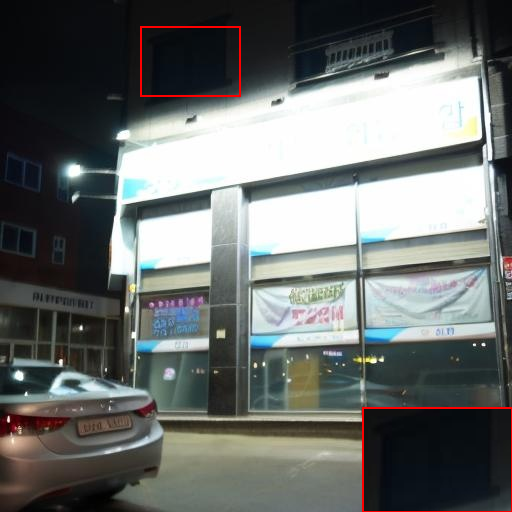}&\includegraphics[height=1.37cm, width=1.9cm]{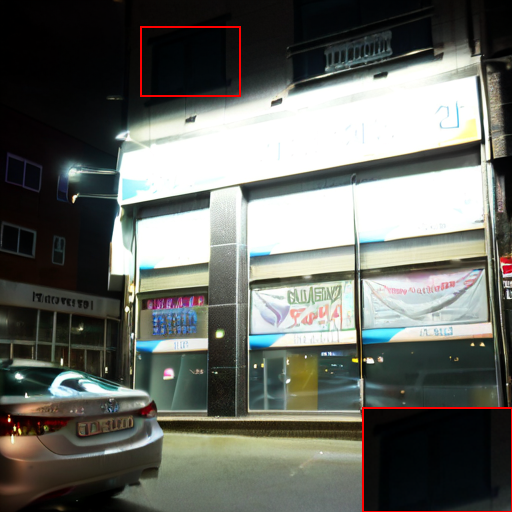}&\includegraphics[height=1.37cm, width=1.9cm]{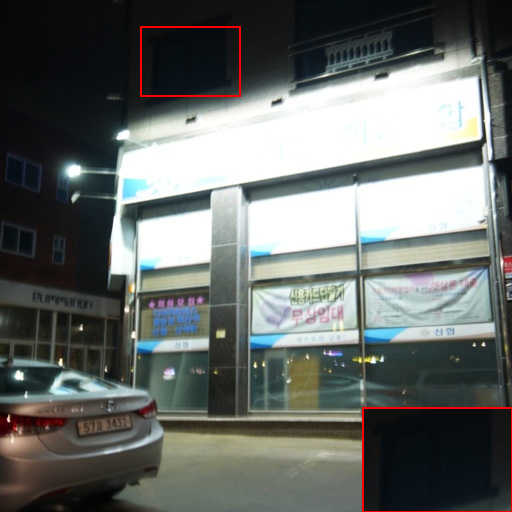}&\includegraphics[height=1.37cm, width=1.9cm]{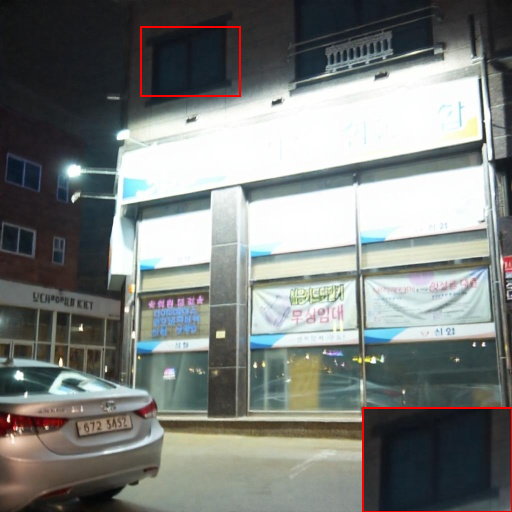}  \\


         \rotatebox[origin=c]{90}{WBSnow\hspace{-32pt}}&\includegraphics[height=1.37cm, width=1.9cm]{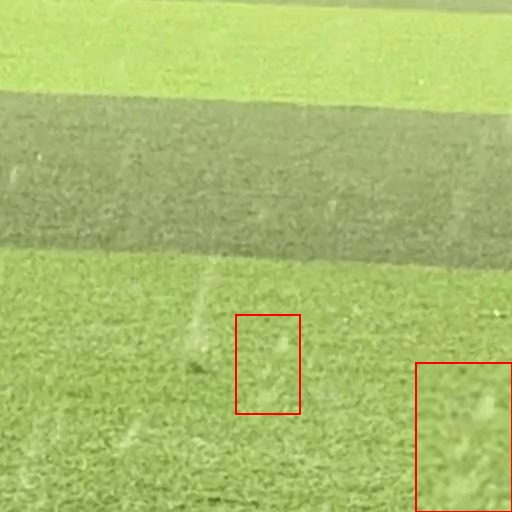}&\includegraphics[height=1.37cm, width=1.9cm]{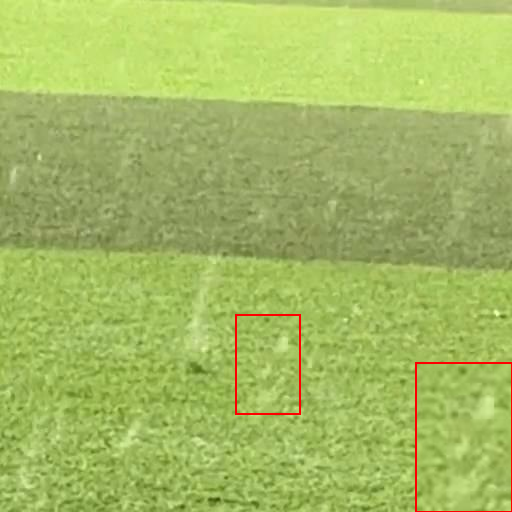}&\includegraphics[height=1.37cm, width=1.9cm]{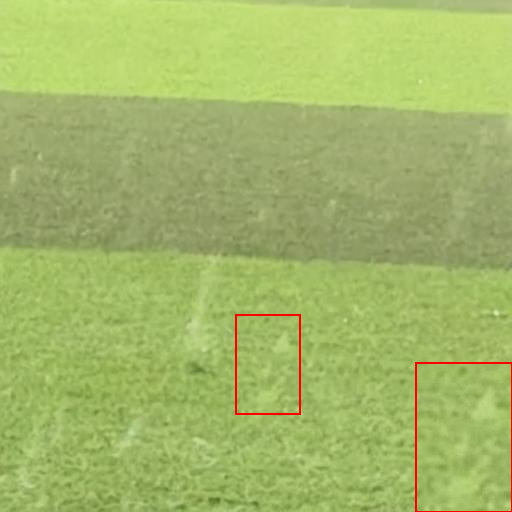}&\includegraphics[height=1.37cm, width=1.9cm]{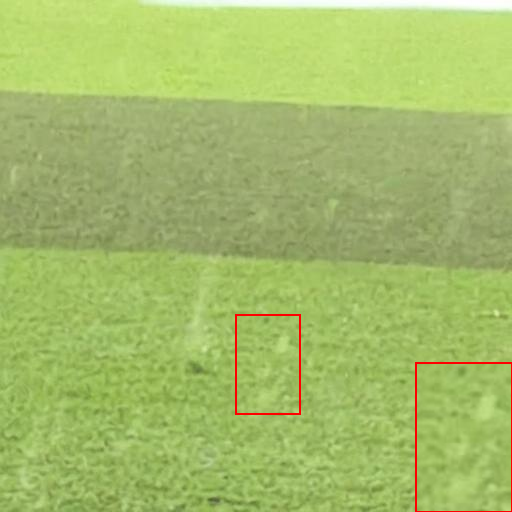}&\includegraphics[height=1.37cm, width=1.9cm]{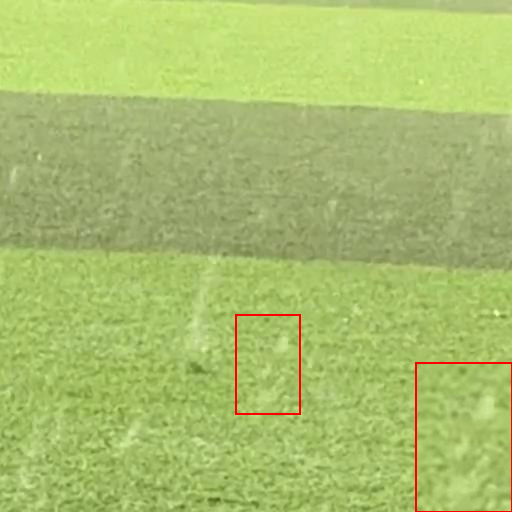}&\includegraphics[height=1.37cm, width=1.9cm]{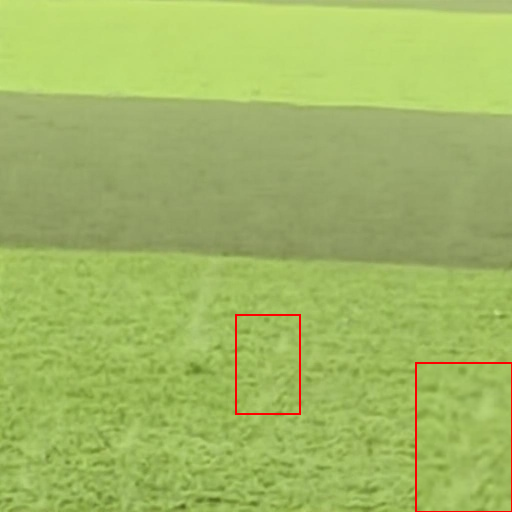}  \\

         \rotatebox[origin=c]{90}{HazeRD\hspace{-32pt}}&\includegraphics[height=1.37cm, width=1.9cm]{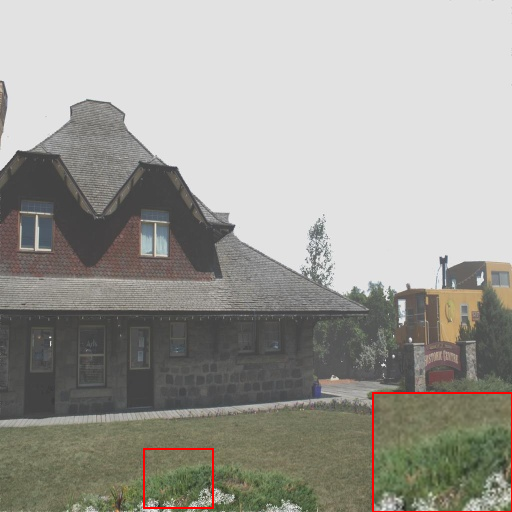}&\includegraphics[height=1.37cm, width=1.9cm]{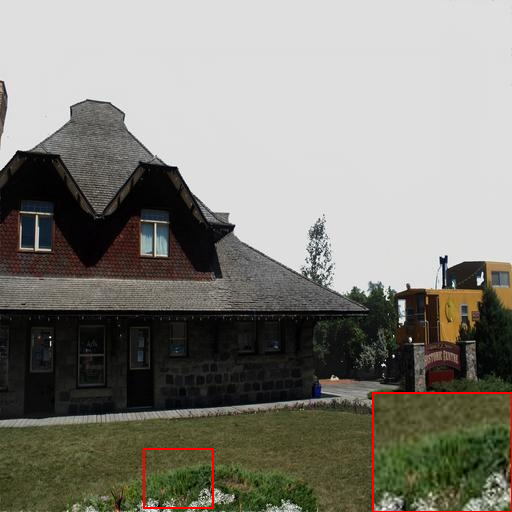}&\includegraphics[height=1.37cm, width=1.9cm]{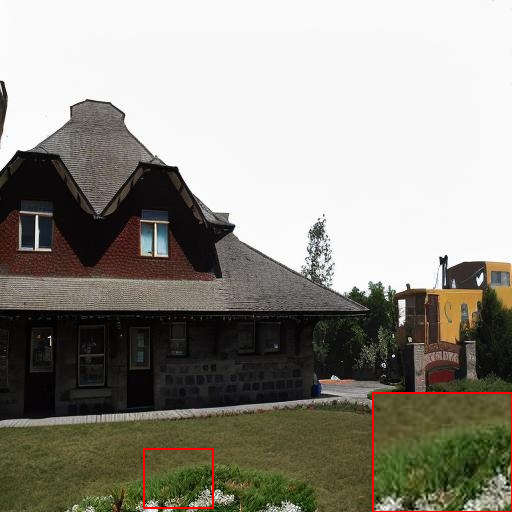}&\includegraphics[height=1.37cm, width=1.9cm]{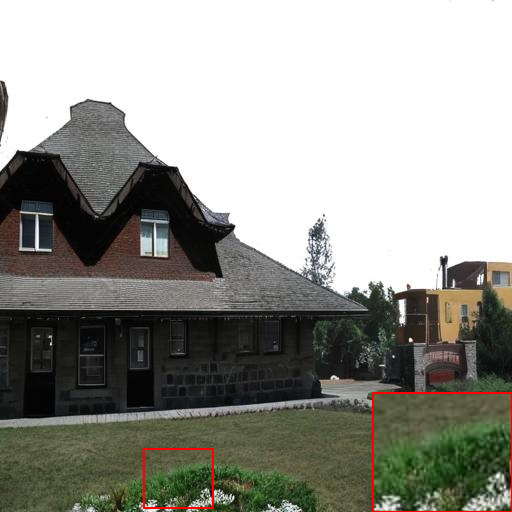}&\includegraphics[height=1.37cm, width=1.9cm]{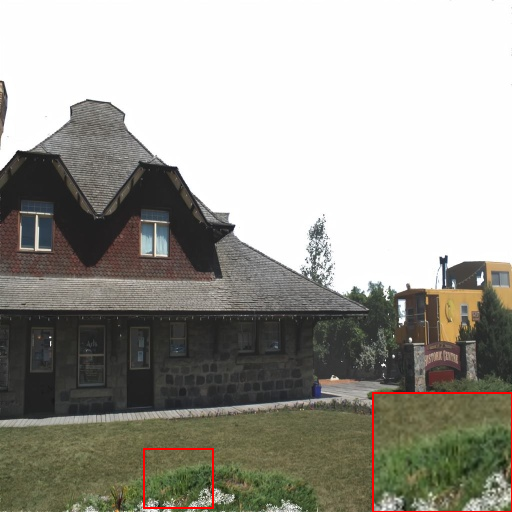}&\includegraphics[height=1.37cm, width=1.9cm]{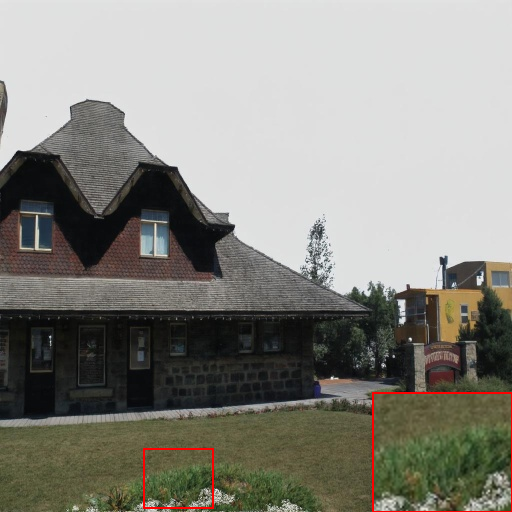}  \\
    \end{tabular}
    \vspace{-5pt}
    \caption{Qualitative comparisons of the pre-trained FLUX model using our learned prompts with state-of-the-art AiOR approaches. Our approach enables the pre-trained FLUX to achieve remarkable restoration performance. WBSnow denotes the snow subset of the WeatherBench~\cite{weatherbench} dataset.}
    \vspace{-5pt}
    \label{fig: qual_img}
\end{figure*}

\begin{figure*}[t]
    \centering
    \small
    \setlength{\tabcolsep}{1pt}
    \begin{tabular}{ccccccc}
    &Input&AutoDIR&FoundIR&ViWS-Net&AverNet&WAN (Ours)\\
          \rotatebox[origin=c]{90}{AAU\hspace{-32pt}}&\includegraphics[height=1.37cm, width=1.9cm]{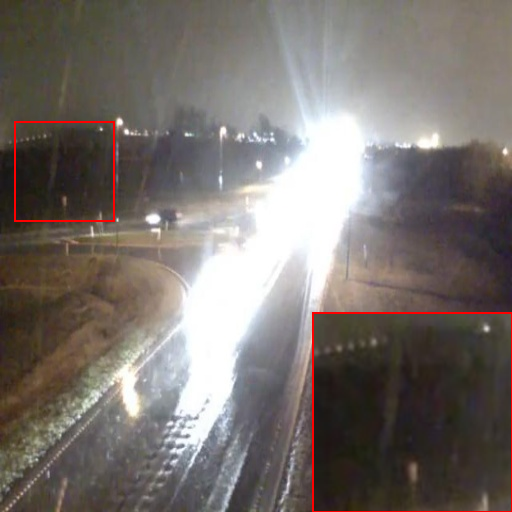}&\includegraphics[height=1.37cm, width=1.9cm]{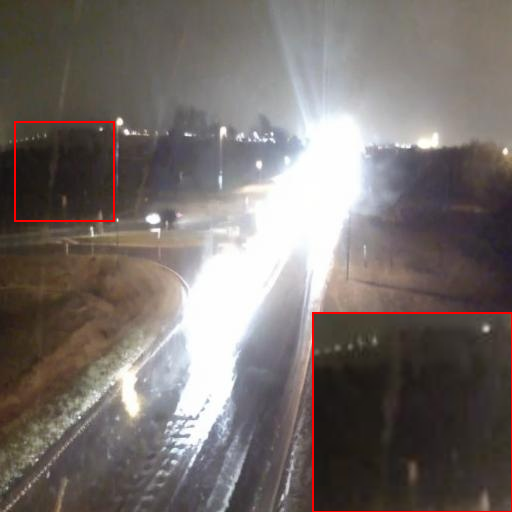}&\includegraphics[height=1.37cm, width=1.9cm]{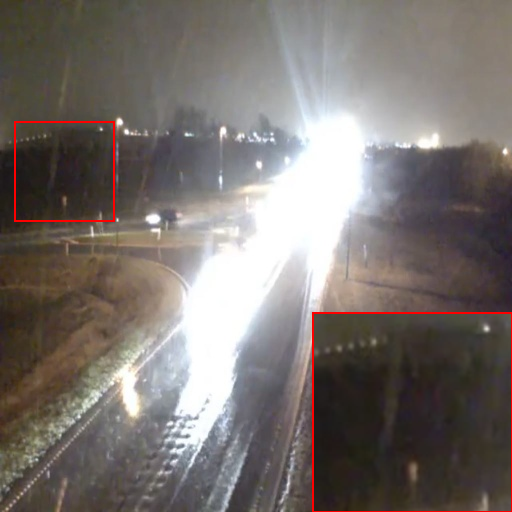}&\includegraphics[height=1.37cm, width=1.9cm]{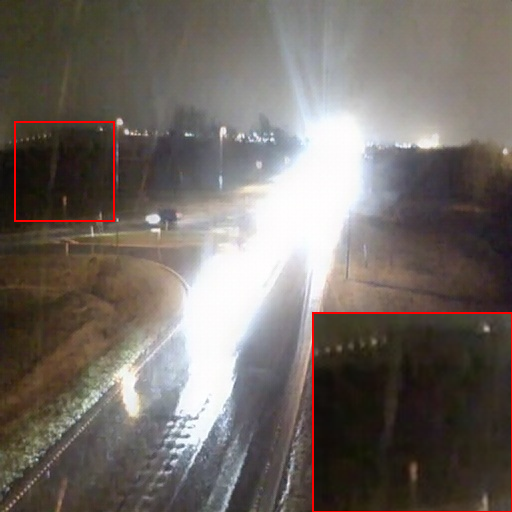}&\includegraphics[height=1.37cm, width=1.9cm]{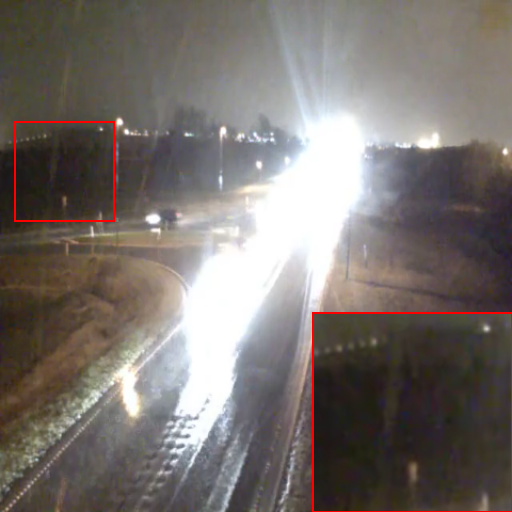}&\includegraphics[height=1.37cm, width=1.9cm]{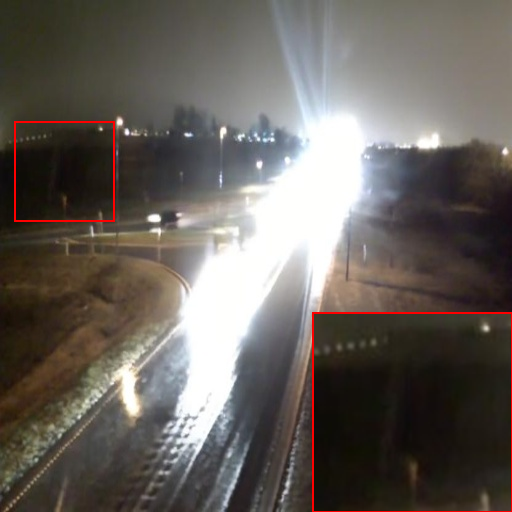}\\

          \rotatebox[origin=c]{90}{NTU\hspace{-32pt}}&\includegraphics[height=1.37cm, width=1.9cm]{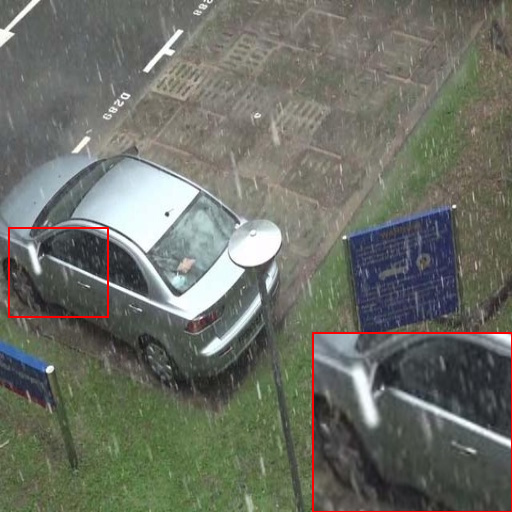}&\includegraphics[height=1.37cm, width=1.9cm]{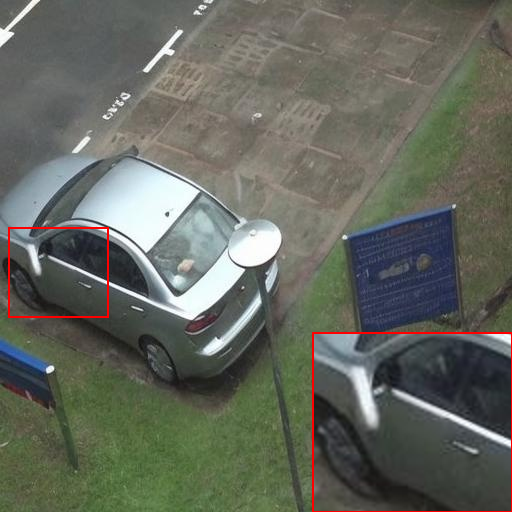}&\includegraphics[height=1.37cm, width=1.9cm]{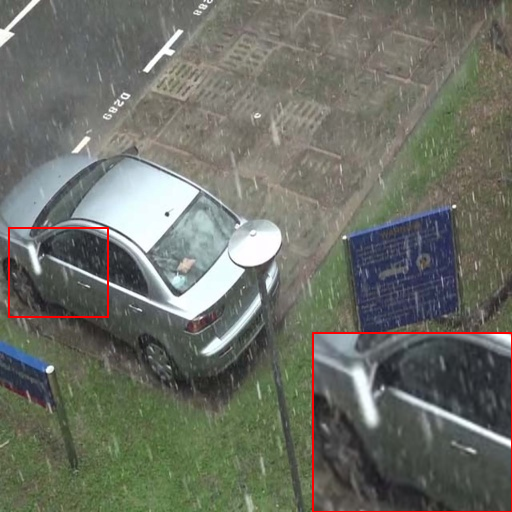}&\includegraphics[height=1.37cm, width=1.9cm]{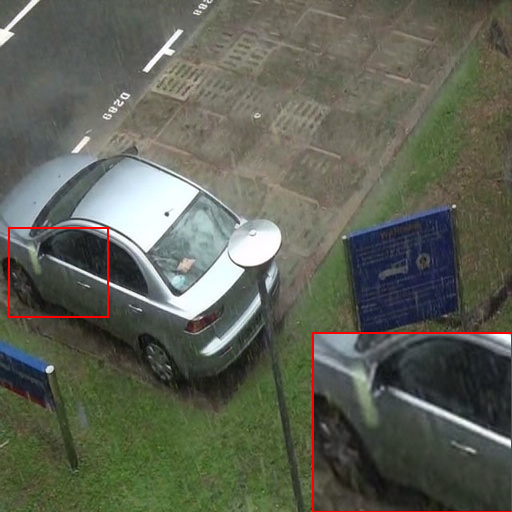}&\includegraphics[height=1.37cm, width=1.9cm]{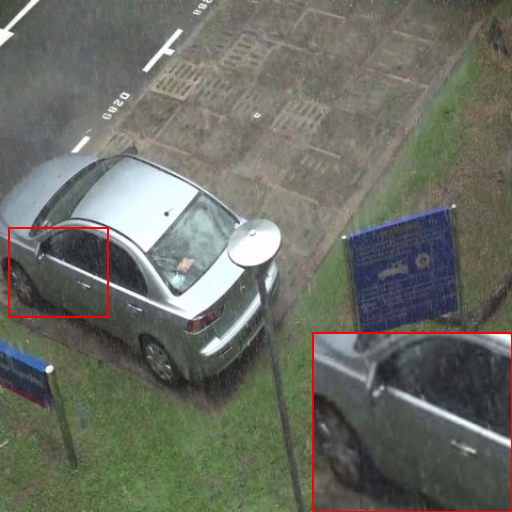}&\includegraphics[height=1.37cm, width=1.9cm]{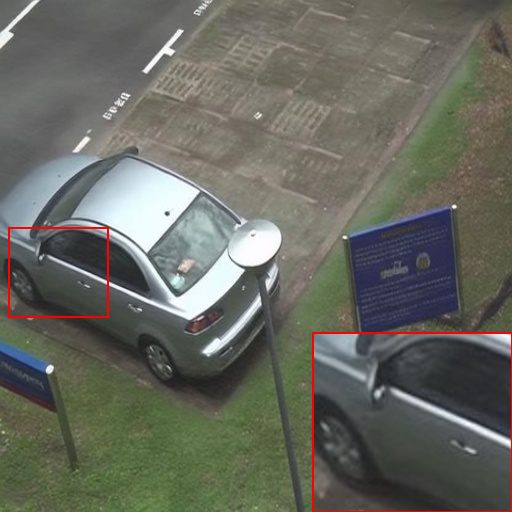}\\

          \rotatebox[origin=c]{90}{RHVD\hspace{-32pt}}&\includegraphics[height=1.37cm, width=1.9cm]{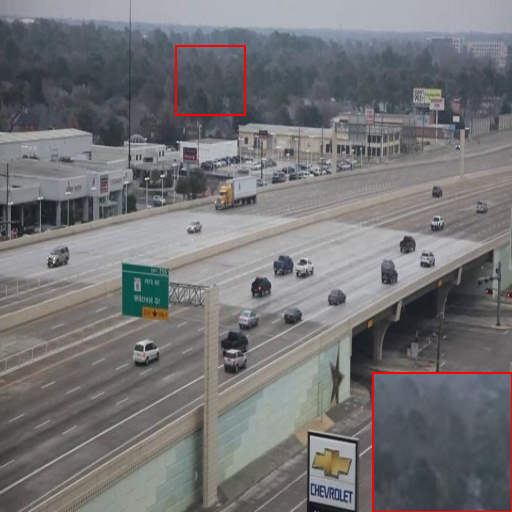}&\includegraphics[height=1.37cm, width=1.9cm]{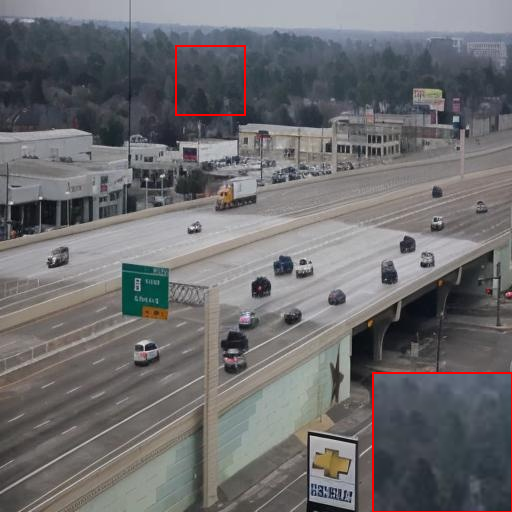}&\includegraphics[height=1.37cm, width=1.9cm]{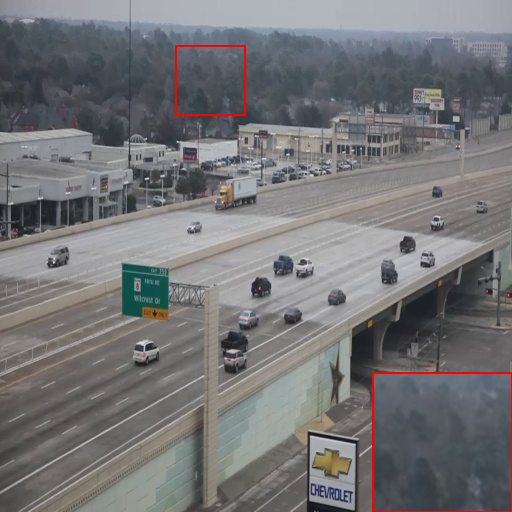}&\includegraphics[height=1.37cm, width=1.9cm]{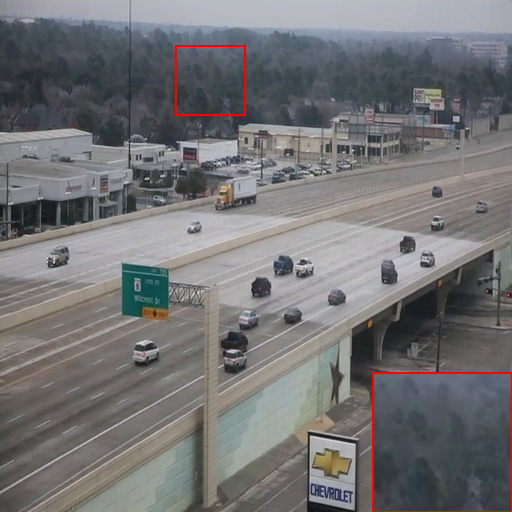}&\includegraphics[height=1.37cm, width=1.9cm]{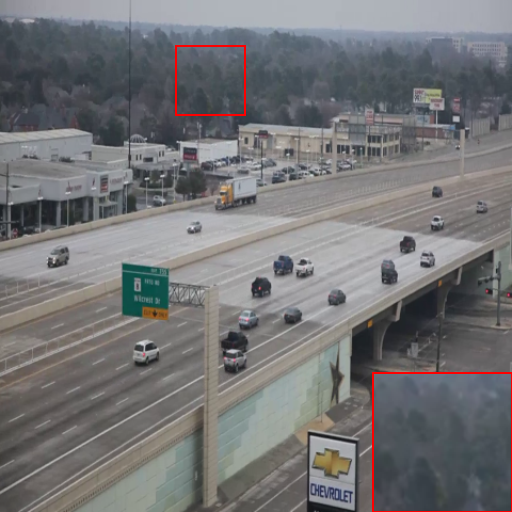}&\includegraphics[height=1.37cm, width=1.9cm]{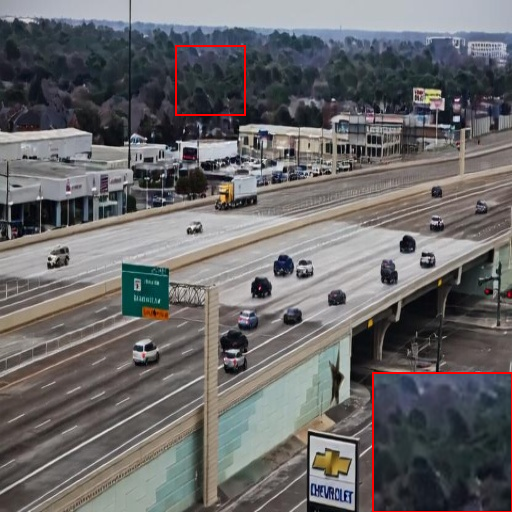}\\

          \rotatebox[origin=c]{90}{4KRD\hspace{-32pt}}&\includegraphics[height=1.37cm, width=1.9cm]{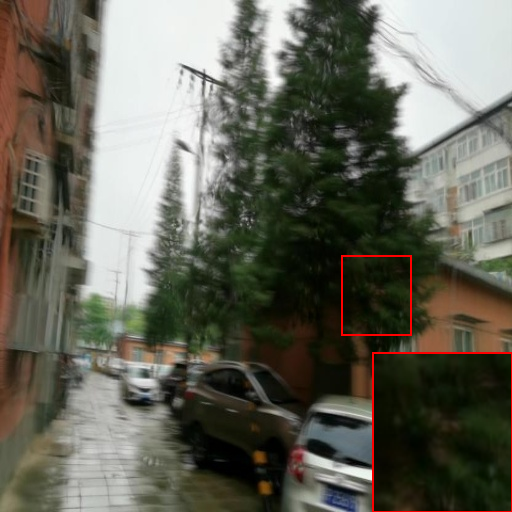}&\includegraphics[height=1.37cm, width=1.9cm]{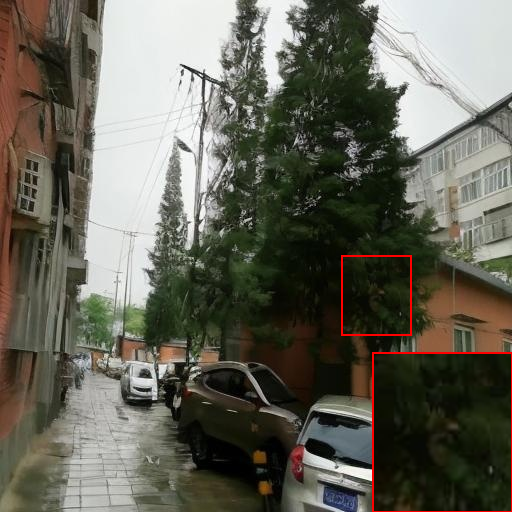}&\includegraphics[height=1.37cm, width=1.9cm]{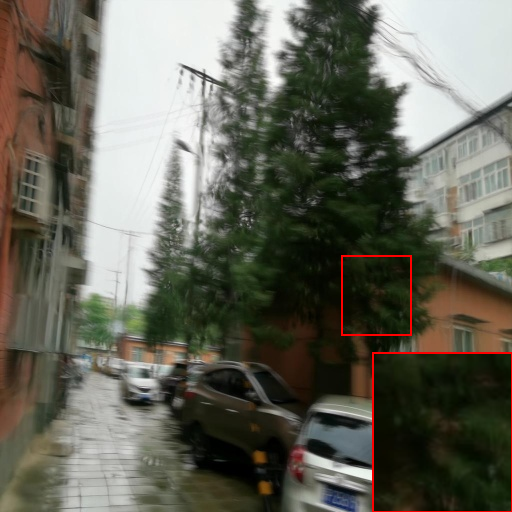}&\includegraphics[height=1.37cm, width=1.9cm]{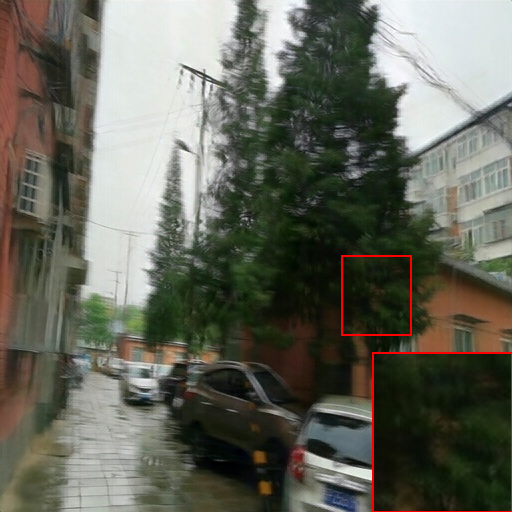}&\includegraphics[height=1.37cm, width=1.9cm]{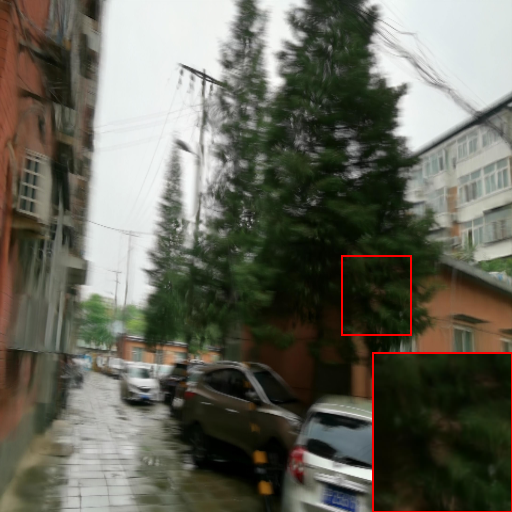}&\includegraphics[height=1.37cm, width=1.9cm]{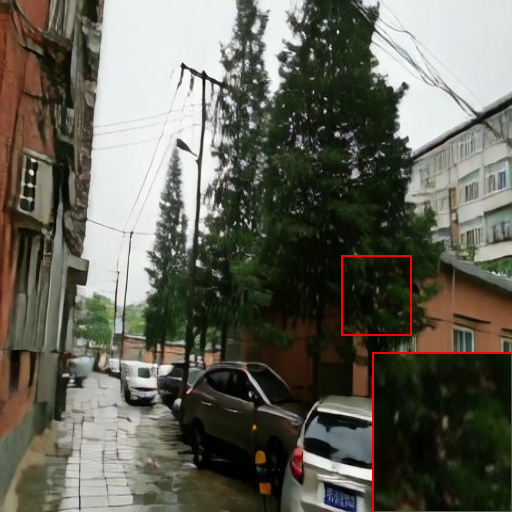}\\

          \rotatebox[origin=c]{90}{LoL-iPhone\hspace{-32pt}}&\includegraphics[height=1.37cm, width=1.9cm]{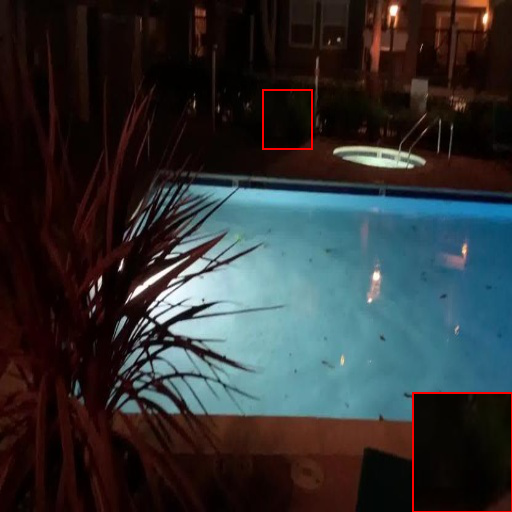}&\includegraphics[height=1.37cm, width=1.9cm]{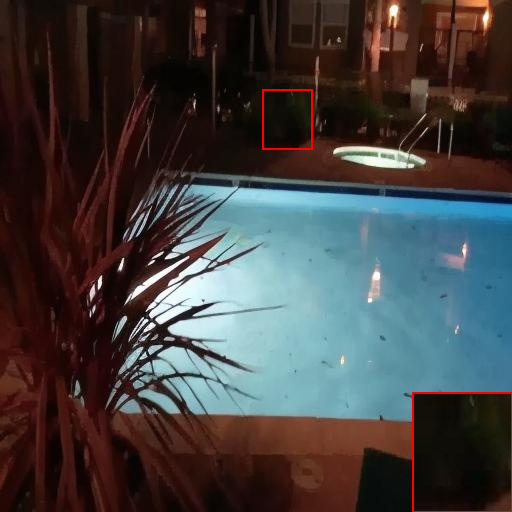}&\includegraphics[height=1.37cm, width=1.9cm]{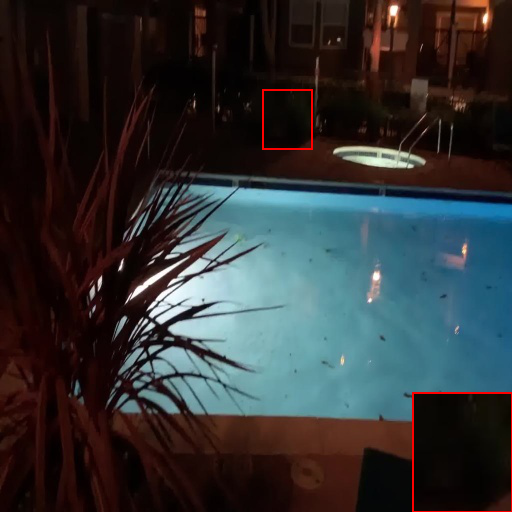}&\includegraphics[height=1.37cm, width=1.9cm]{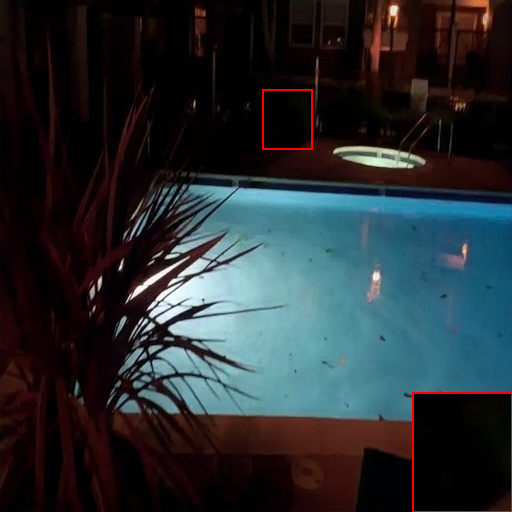}&\includegraphics[height=1.37cm, width=1.9cm]{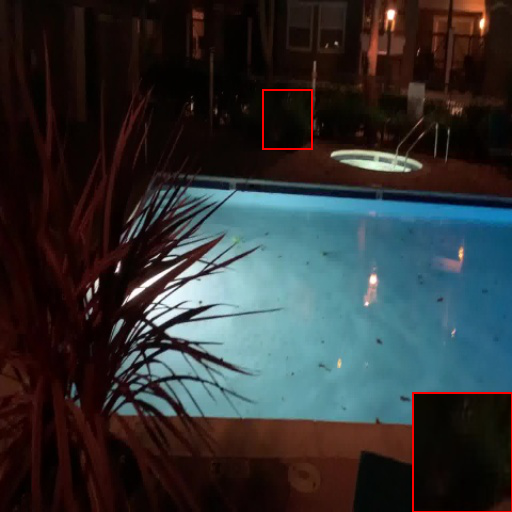}&\includegraphics[height=1.37cm, width=1.9cm]{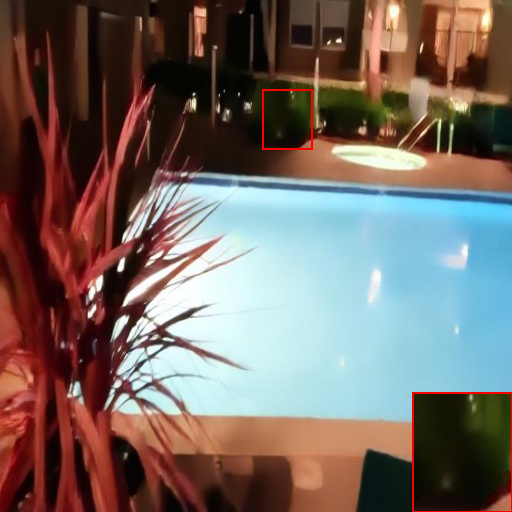}\\
    \end{tabular}
    \vspace{-5pt}
    \caption{Qualitative comparisons of the pre-trained WAN model using our learned prompts with state-of-the-art AiOR approaches. ViWS-Net and AverNet are video restoration approaches while others are proposed for image restoration. Our prompts elicit the strong restoration potential of the pre-trained WAN model. AAU: AAURainSnow~\cite{aau}, NTU: real test set of NTU-Rain~\cite{spac}.}
    \label{fig: qual_vid}
\end{figure*}

\subsection{Comparisons}
\label{subsec: comparisons}

\noindent\textbf{Image restoration.}
We instantiate our learned prompts on the \emph{pre-trained} FLUX backbone and compare against recent all-in-one restoration (AiOR) methods, including DCPT~\cite{dcpt}, DFPIR~\cite{dfpir}, AutoDIR~\cite{autodir}, PixWizard~\cite{pixwizard}, and FoundIR~\cite{foundir}, using their publicly available checkpoints.
We report both distortion and perceptual metrics: PSNR (P) in dB, SSIM (S), LPIPS (L)~\cite{lpips}, DISTS (D)~\cite{dists}, CLIPIQA (C)~\cite{clipiqa}, and MUSIQ (M)~\cite{musiq}. The mean within-distribution (WD) performance for all methods is reported in Table~\ref{tab: quant_wd}. We note that methods based on latent diffusion models are impacted by the reconstruction bottleneck of the VAE, which can affect PSNR/SSIM scores even when the restoration is of high quality. This has also been observed in other works~\cite{pixwizard,restorevar}. More detailed WD results are given in the supplementary.

Testing on out-of-distribution (OOD), mixed, and unseen degradations provides a more faithful measure of performance under real-world and complex corruptions. As shown in Table~\ref{tab:quant_img_ood}, our method remains consistently competitive across all three settings, often delivering strong perceptual scores. For the unseen TOLED and POLED datasets, we condition using the low-light and blur prompts, as they were a close visual match to the under-display camera degradations.

\noindent\textbf{Video restoration.}
For videos, we compare our WAN-based instantiation to ViWS-Net~\cite{viwsnet} and AverNet~\cite{avernet}, and additionally report DOVER (Do)~\cite{dover} as a video quality metric. We also report the scores for image-based approaches. 
Since the publicly available checkpoints of AverNet and ViWS-Net did not address many of our restoration tasks, we re-trained them on the same training data. WD results are reported in Table~\ref{tab: quant_wd} and comparisons on OOD data is given in Table~\ref{tab:quant_video_ood}. As for images, we provide detailed WD results in the supplementary. Table~\ref{tab:quant_video_ood} shows that our prompt-tuned WAN achieves strong performance for the OOD degradations. Moreover, it outperforms image and video restoration performance on the DOVER video quality assessment metric as the pre-trained WAN has very strong temporal priors.

\noindent\textbf{Qualitative evaluation.}
Beyond metrics, we visualize results on OOD, mixed, and unseen degradations in Fig.~\ref{fig: qual_img} for image restoration datasets and Fig.~\ref{fig: qual_vid} for video restoration datasets. In the interest of space, we only provide qualitative comparisons for specific OOD, mixed and unseen degradation datasets. Additional results on the remaining OOD datasets, within-distribution benchmarks, and comparisons with DCPT are provided in the supplementary material.

While several methods achieve strong scores on specific datasets, we observe that they often fail to restore under such complex conditions. For example, in Fig.~\ref{fig: qual_img}, our prompt-instantiated FLUX model is able to achieve better visual restoration than the state-of-the-art approaches. A similar observation can be inferred from Fig.~\ref{fig: qual_vid}. To facilitate a clearer comparison against our prompt-instantiated WAN model for video restoration, we include video clips in the supplementary material.

\begin{figure*}[t]
  \centering

  \begin{minipage}[t]{0.24\textwidth}\centering\small Input \end{minipage}\hfill
  \begin{minipage}[t]{0.24\textwidth}\centering\small Naive training\end{minipage}\hfill
  \begin{minipage}[t]{0.24\textwidth}\centering\small DDBM+Prompt\end{minipage}\hfill
  \begin{minipage}[t]{0.24\textwidth}\centering\small EBR+Prompt \end{minipage}

  \begin{subfigure}[t]{0.24\textwidth}
    \centering
    \includegraphics[width=\linewidth]{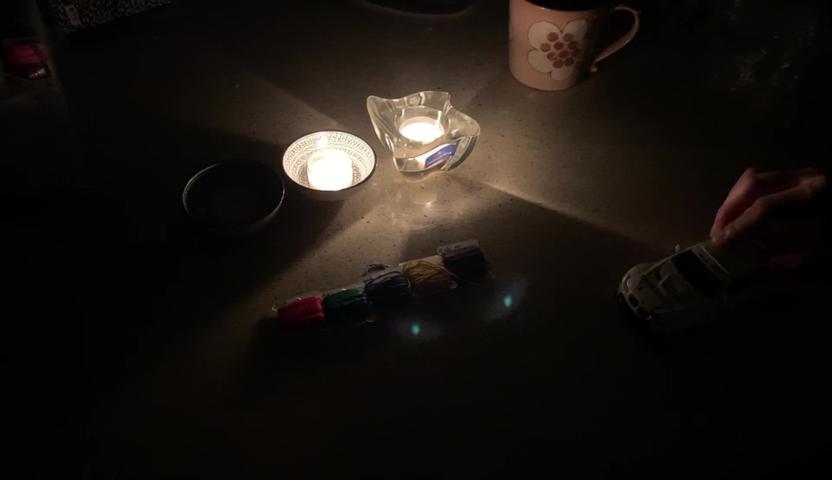}
  \end{subfigure}\hfill
  \begin{subfigure}[t]{0.24\textwidth}
    \centering
    \includegraphics[width=\linewidth]{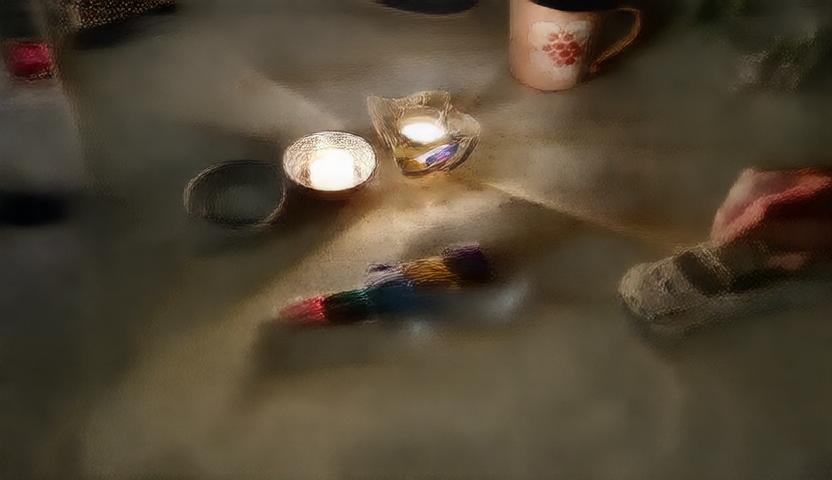}
  \end{subfigure}\hfill
  \begin{subfigure}[t]{0.24\textwidth}
    \centering
    \includegraphics[width=\linewidth]{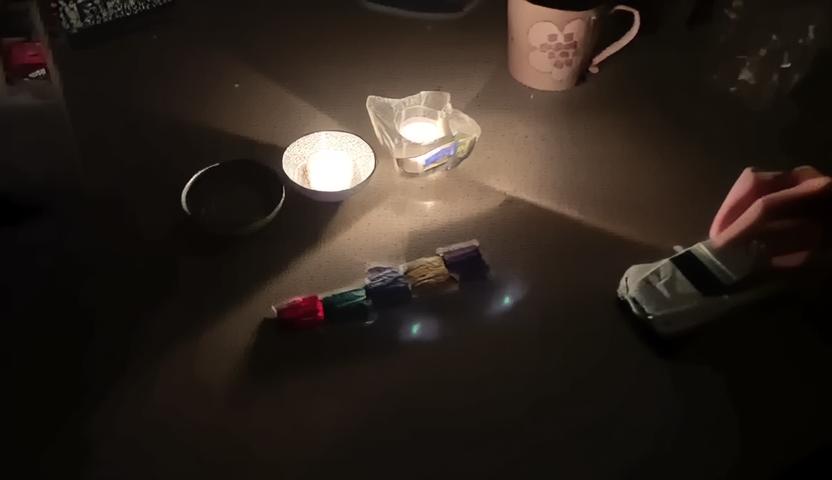}
  \end{subfigure}\hfill
  \begin{subfigure}[t]{0.24\textwidth}
    \centering
    \includegraphics[width=\linewidth]{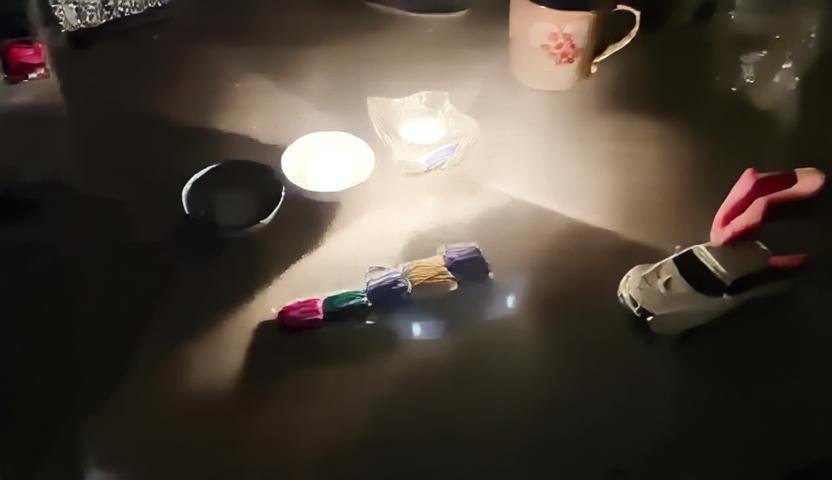}
  \end{subfigure}

    \vspace{-5pt}
  \caption{Although naive prompt training enhances the image, it produces several artifacts due to trajectory mismatch. DDBM with prompt enhances the image marginally. EBR provides best results.}
  \vspace{-10pt}
  \label{fig: bridge_ablt}
\end{figure*}





\begin{figure}[t]
\centering

\begin{minipage}[t]{0.45\linewidth}
  \vspace{10pt}
  \centering
  \includegraphics[width=\linewidth]{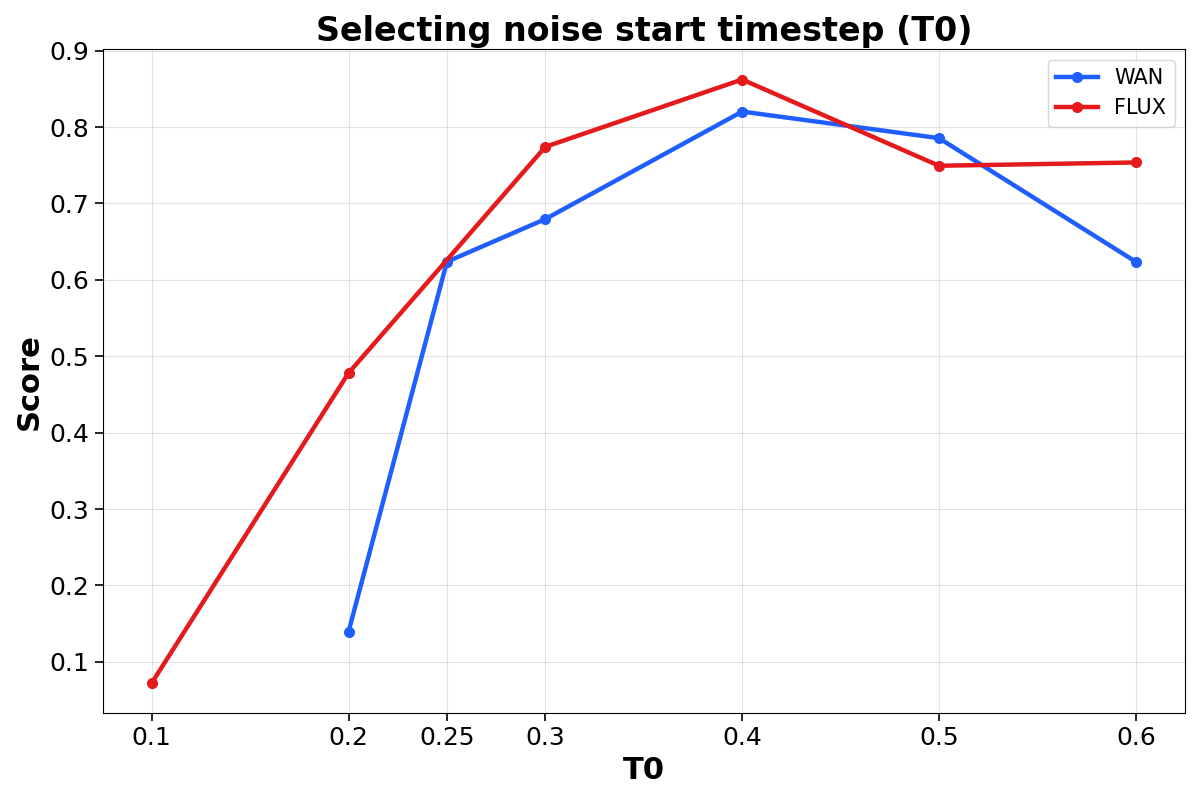}
  \caption{Plot of normalized scores (Sec.~\ref{subsec: ablation}) for determining the best value of $T_0$ across different candidates: $0.1,0.2,0.3,0.4,0.5,0.6$. A higher score is better.}
  \label{fig:t0}
\end{minipage}\hfill
\begin{minipage}[t]{0.54\linewidth}
  \vspace{0pt}
  \centering
  \scriptsize

  \captionsetup{type=table}
  \caption{Quantitative comparison of naive prompt tuning, prompt tuning with DDBM-like~\cite{ddbm} bridge and with EBR-like~\cite{gcb} bridge (Ours).}
  \label{tab:abl_bridge_wan}
  \setlength{\tabcolsep}{3pt}
  \adjustbox{max width=\linewidth}{%
  \begin{tabular}{l|cccccc}
    \toprule
    \textbf{Training trajectory} & \textbf{P$\uparrow$} & \textbf{S$\uparrow$} & \textbf{L$\downarrow$} & \textbf{D$\downarrow$} & \textbf{C$\uparrow$} & \textbf{M$\uparrow$} \\
    \midrule
    Naive        & 21.17&0.694&0.228&0.183&0.474&49.10 \\
    DDBM bridge  & 21.44&0.712&0.200&0.121&0.497&55.64 \\
    EBR Bridge   & 22.49&0.728&0.162&0.108&0.529&60.60 \\
    \bottomrule
  \end{tabular}}

  \vspace{2pt} 

  \captionsetup{type=table}
  \caption{Ablation on prompt learning for WAN: Learning prompts in the text-encoder embedding space vs.\ a residual prompt injected in the attention context space.}
  \label{tab:res_prompt}
  \setlength{\tabcolsep}{4pt}
  \adjustbox{max width=\linewidth}{%
  \begin{tabular}{l|cccccc}
    \toprule
    \textbf{Conditioning} & \textbf{P$\uparrow$} & \textbf{S$\uparrow$} & \textbf{L$\downarrow$} & \textbf{D$\downarrow$} & \textbf{C$\uparrow$} & \textbf{M$\uparrow$} \\
    \midrule
    Embedding-space & 20.30 & 0.753 & 0.213 & 0.151 & 0.538 & 64.87 \\
    Residual prompt & 20.86 & 0.770 & 0.206 & 0.152 & 0.540 & 65.00 \\
    \bottomrule
  \end{tabular}}
\end{minipage}

\end{figure}
\vspace{-10pt}

\subsection{Ablations}
\label{subsec: ablation}

We now ablate key components of our framework.

\noindent\textbf{Selecting the start noise level $T_0$.}
$T_0$ controls the amount of noise in the degraded endpoint of the bridge (Eq.~\ref{eq:gcb_final}).
We tune $T_0 \in \{0.1,0.2,0.3,0.4,0.5,0.6\}$ on the \emph{most severe} degradation in our training set, i.e. low-light (see Sec.~\ref{subsec: design_choices}).
To make selection robust across distortion and perceptual criteria, we aggregate multiple metrics (PSNR, SSIM, LPIPS, DISTS, CLIPIQA, MUSIQ) into a single score by (i) min-max normalizing each metric across candidate $T_0$ values and (ii) averaging the normalized scores with equal weights (full procedure in the supplementary). As shown in Fig.~\ref{fig:t0}, $T_0=0.4$ achieves the best overall score for both WAN and FLUX.

\noindent\textbf{Bridge resolves trajectory mismatch.}
We compare three training/sampling state constructions for prompt learning:
(i) Naive prompt tuning on states anchored at $z_{\text{deg}}$ (Eq.~\ref{eq:deg_forward_fm}),
(ii) prompt tuning using a DDBM-style bridge~\cite{ddbm} (Eqn.~\ref{eq:ddbm_reparam2}),
and (iii) prompt tuning using the monotone noisy-degraded $\rightarrow$ clean bridge (EBR-style~\cite{gcb}; Eqn.~\ref{eq:desired_bridge}/\ref{eq:gcb_final}).
Table~\ref{tab:abl_bridge_wan} shows that naive tuning yields lowest performance, while using a DDBM-style bridge improves performance. The EBR-style bridge performs best. Qualitative comparisons in Fig.~\ref{fig: bridge_ablt} show that naive prompt tuning produces undesirable artifacts while the DDBM-style bridge under-corrects.

\noindent\textbf{Residual prompt injection at attention.}
Instead of directly learning the full context in text-encoder output space, we learn a residual prompt added to the null-text context at the attention input, with a learned gate initialized at zero.
This design improves performance and reduces the parameter budget compared to directly optimizing text-encoder outputs.
On WAN, this approach improves performance over direct context embedding learning (Table~\ref{tab:res_prompt}), while using fewer parameters (e.g., $\sim$300K vs.\ $\sim$900K).


We provide more ablations in the supplementary along with a discussion of the limitations of our work.

\section{Conclusion}
\label{sec: conclusion}

In this work, we showed that pre-trained diffusion models possess restoration behavior that can be unlocked by training lightweight prompts. Our key finding is that this restoration behavior is largely inaccessible through natural-language prompting or token-space prompt optimization but can be elicited by directly learning context embeddings in the text-encoder output space. We further showed that naive prompt learning can be unstable due to a train-test trajectory mismatch between the degraded image noising process used for prompt optimization and the states visited during reverse-time sampling. To address this, we trained prompts on bridge-defined intermediate states that align training and inference dynamics, yielding a coherent denoising path from noisy degraded inputs to clean outputs. Building on these insights, we adapt the FLUX image model and WAN video model using only lightweight learned prompts, without fine-tuning or restoration-specific control modules. Comprehensive evaluations show that our approach delivers strong generalization and competitive performance to state-of-the-art approaches.

\section*{Acknowledgments}
This work is supported by the Intelligence Advanced Research Projects Activity (IARPA) via Department of Interior/ Interior Business Center (DOI/IBC) contract number 140D0423C0076. The U.S. Government is authorized to reproduce and distribute reprints for Governmental purposes notwithstanding any copyright annotation thereon. Disclaimer: The views and conclusions contained herein are those of the authors and should not be interpreted as necessarily representing the official policies or endorsements, either expressed or implied, of IARPA, DOI/IBC, or the U.S. Government.

%
%
\bibliographystyle{splncs04}
\bibliography{main}
\end{document}